\documentclass{article}

\usepackage{microtype}
\usepackage{graphicx}
\usepackage{subfigure}
\usepackage{booktabs} 

\usepackage{hyperref}

\usepackage[accepted]{icml2020}
\usepackage[utf8]{inputenc} 
\usepackage[T1]{fontenc}    
\usepackage{url}            
\usepackage{amsfonts}       
\usepackage{nicefrac}       
\usepackage{microtype}      
\usepackage{import}

\usepackage{makecell, cellspace}

\usepackage[page]{appendix}

\usepackage{amsthm}
\usepackage{amsmath,amssymb}
\usepackage{amsfonts}
\usepackage{bbding}

\usepackage{mathtools}
\usepackage{algorithm}
\usepackage{algorithmic}
\usepackage{tikz}
\usepackage{multirow}
\usetikzlibrary{shapes.geometric, arrows, automata, positioning}
\usepackage{makecell}

\usepackage{dsfont}
\usepackage{enumerate}   
\usepackage{natbib}

\DeclarePairedDelimiter\br{(}{)}
\DeclarePairedDelimiter\brs{[}{]}
\DeclarePairedDelimiter\brc{\{}{\}}
\DeclarePairedDelimiter\inner{\langle}{\rangle}
\DeclarePairedDelimiter\abs{\lvert}{\rvert}

\usepackage{verbatim}

\usepackage{bbm}

\DeclareMathOperator*{\argmin}{arg\,min}
\DeclareMathOperator{\aset}{\mathcal{A}}
\DeclareMathOperator{\sset}{\mathcal{S}}

\def\S{{\mathcal S}}
\def\A{{\mathcal A}}
\def\F{{\mathcal F}}
\def\P{{\mathcal P}}

\def\Regret{\mathrm{Regret}}

\def\filt{\mathcal{F}_{k-1}}

\makeatletter
\newcommand*{\rom}[1]{\expandafter\@slowromancap\romannumeral #1@}
\makeatother

\def\E{{\mathbb E}}

\def\M{{\mathcal M}}

\def\I{{\mathbb I}}

\def\R{\mathbb R}

\def\triangleq{:=}

\newcommand*\dkl[2]{d_{KL}(#1||#2)}
\newcommand*\bregman[2]{B_\omega\left(#1,#2\right)}

\usepackage{bbm}
\newcommand{\ind}{\mathbbm{1}}

\DeclareMathOperator{\simplex}{\Delta_{\A}}

\newcommand*\norm[1]{\left\|#1\right\|}

\usepackage{array}
\newcolumntype{P}[1]{>{\centering\arraybackslash}p{#1}}
\newcolumntype{L}[1]{>{\raggedright\arraybackslash}p{#1}}

\usepackage{thmtools}
\usepackage{thm-restate}

\newtheorem{lemma}{Lemma}
\newtheorem{corollary}{Corollary}

\usepackage{todonotes}

\newtheorem{claim}{Claim}

\newtheorem{remark}{Remark}
\numberwithin{equation}{section}
\numberwithin{remark}{section}

\newenvironment{proofsketch}{%
  \proof}{\endproof}

\definecolor{darkgray}{rgb}{0.66, 0.66, 0.66}

\renewcommand{\appendixtocname}{List of Appendices}

\makeatletter
\let\oldappendix\appendices

\renewcommand{\appendices}{%
  \clearpage
  \renewcommand{\thesection}{\Alph{section}}
  \let\tf@toc\tf@app
  \addtocontents{app}{\protect\setcounter{tocdepth}{2}}
  \immediate\write\@auxout{%
    \string\let\string\tf@toc\string\tf@app^^J
  }
  \oldappendix
}%

\newcommand{\listofappendices}{%
  \begingroup
  \renewcommand{\contentsname}{\appendixtocname}
  \let\@oldstarttoc\@starttoc
  \def\@starttoc##1{\@oldstarttoc{app}}
  \tableofcontents
  \endgroup
}

\makeatother

\usepackage[symbol]{footmisc}

\icmltitlerunning{Optimistic Policy Optimization with Bandit Feedback
}

\renewcommand{\algorithmiccomment}[1]{\textit{#1}}

\begin{document}

\twocolumn[
\icmltitle{Optimistic Policy Optimization with Bandit Feedback}



\icmlsetsymbol{equal}{*}

\begin{icmlauthorlist}
\icmlauthor{Yonathan Efroni}{equal,tech}
\icmlauthor{Lior Shani}{equal,tech}
\icmlauthor{Aviv Rosenberg}{tau}
\icmlauthor{Shie Mannor}{tech}
\end{icmlauthorlist}

\icmlaffiliation{tech}{Technion - Israel Institute of Technology, Haifa, Israel}
\icmlaffiliation{tau}{Tel Aviv University, Tel Aviv, Israel}

\icmlcorrespondingauthor{Lior Shani}{shanlior@gmail.com}
\icmlcorrespondingauthor{Yonathan Efroni}{jonathan.efroni@gmail.com}

\icmlkeywords{Machine Learning, Reinforcement Learning, Optimization, Mirror Descent}

\vskip 0.3in
]



\printAffiliationsAndNotice{\icmlEqualContribution} 

\begin{abstract}
    Policy optimization methods are one of the most widely used classes of Reinforcement Learning (RL) algorithms. Yet, so far, such methods have been mostly analyzed from an optimization perspective, without addressing the problem of exploration, or by making strong assumptions on the interaction with the environment. 
    In this paper we consider model-based RL in the tabular finite-horizon MDP setting with unknown transitions and bandit feedback. For this setting, we propose an optimistic policy optimization algorithm for which we establish $\tilde O(\sqrt{S^2 A H^4 K})$ regret for stochastic rewards. Furthermore, we prove $\tilde O( \sqrt{ S^2 A H^4 }  K^{2/3} ) $ regret for adversarial rewards. Interestingly, this result matches previous bounds derived for the bandit feedback case, yet with known transitions. To the best of our knowledge, the two results are the first sub-linear regret bounds obtained for policy optimization algorithms with unknown transitions and bandit feedback.
\end{abstract}

\section{Introduction}

Policy Optimization (PO) is among the most widely used methods in Reinforcement Learning (RL)~\cite{peters2006policy,peters2008reinforcement,deisenroth2011pilco,lillicrap2015continuous,levine2016end,gu2017deep}. Unlike value-based approaches, e.g., Q-learning, these types of methods directly optimize the policy by incrementally changing it. Furthermore, PO methods span wide variety of popular algorithms such as policy-gradient algorithms~\cite{sutton2000policy}, natural policy gradient~\cite{kakade2002natural}, trust region policy optimization (TRPO)~\cite{schulman2015trust} and soft actor-critic~\cite{haarnoja2018soft}.  

Due to their popularity, there is a rich  literature that provides different types of theoretical guarantees for different PO methods~\cite{scherrer2014local,abbasi2019politex,agarwal2019optimality,liu2019neural,bhandari2019global,shani2019adaptive,wei2019model} for both the approximate and tabular settings. However, previous results, concerned with regret or PAC bounds for the RL setting when the model is unknown and only bandit feedback is given, provide guarantees which critically depend on `concentrability coefficients'~\cite{kakade2002approximately,munos2003error,scherrer2014approximate} or on a unichain MDP assumption \cite{abbasi2019politex}. However, these coefficients might be infinite and are usually small only for highly stochastic domains, while the unichain assumption is often very restrictive.

Recently, \citet{cai2019provably} established an $\tilde O(\sqrt{K})$ regret bound for an optimistic PO method in the case of an unknown model and assuming full-information feedback on adversarially chosen instantaneous costs, where $K$ is the number of episodes seen by the agent. In this work, we eliminate the full-information assumption on the cost, as in most practical settings only bandit feedback on the cost is given, i.e., the cost is observed through interacting with the environment. Specifically, we establish regret bounds for an optimistic PO method in the case of an unknown model and bandit feedback on the instantaneous cost in two regimes:
\begin{enumerate}
    \item For stochastic cost, we establish an $\tilde O(\sqrt{S^2 A H^4 K})$ regret bound for a PO method (Section~\ref{sec: stochastic PO}). 
    \item For adversarially chosen cost, we establish an $\tilde O( \sqrt{ S^2 AH^4}  K^{2/3} )$ regret bound for a PO method. The regret bound matches the $\tilde{O}\br*{K^{2/3}}$ upper bound obtained by~\citet{neu2010onlineT23} for PO methods which have an access to the true model and observe bandit adversarial cost feedback (Section~\ref{sec: adveresial po}). 
\end{enumerate}




\begin{table*}
\caption{Comparison of our bounds with several state-of-the-art bounds for policy-based RL and occupancy measure RL in tabular finite-horizon MDPs. The time complexity of the algorithms is per episode; $S$ and $A$ are the sizes of the state and action sets, respectively; $H$ is the horizon of the MDP; $K$ is the total number of episodes; Env. describes the environment of the algorithm: stochastic (Sto) or adversarial (Adv); Policy based describes if an algorithm is based on policy updates or on occupancy measure updates. Costs and model terms describes how optimism is used in the estimators: For costs, a bonus term (Bonus) or an importance sampling estimator (IS). For transition model: a bonus term (Bonus) or a confidence interval of models (CI); The update procedure describes how the optimization problem is solved, using a state-wise closed-form solution (Closed form), or by solving an optimization problem over the entire state-action space (Optimization).
The algorithms proposed in this paper are highlighted in gray. The other algorithms are OMD-BP \cite{neu2010online}, UC-O-REPS \cite{rosenberg2019bandit}, OPPO \cite{cai2019provably} and UOB-REPS \cite{jin2019learning}. (*)~represents the different setting of the average cost criterion.
}
\begin{center}
\begin{tabular}[c]{|c|c|c|c|c|c|c|c|c|}\hline
 Algorithm & Regret & Env. & \makecell{Bandit \\ Feedback} & \makecell{Unknown \\ Model} &\makecell{Policy \\ Based} &  \makecell{ Costs} & \makecell{ Model } &  \makecell{Update \\ Procedure} \\ \hline \hline
      \rowcolor{lightgray} \texttt{POMD} & $\tilde{O}(\sqrt{S^2 A H^4 K})$ & Sto. & \Checkmark
 & \Checkmark & \Checkmark
 & Bonus & Bonus & Closed form \\ 
\hline \hline
  \texttt{OMDP-BP(*)} & $ \tilde{O}(K^{2/3})$  & Adv. & \Checkmark & \XSolidBrush & \Checkmark & IS & - & Closed form\\
\hline
  \texttt{UC-O-REPS} & $ \tilde{O}(\sqrt{S^2 A H^4 K})$ & Adv. & \XSolidBrush & \Checkmark & \XSolidBrush & - & CI & Optimization\\ 
\hline
  \texttt{OPPO} & $\tilde{O}( \sqrt{S^3 A^3 H^4 K})$ & Adv. & \XSolidBrush & \Checkmark & \Checkmark & - & Bonus & Closed form\\ 
\hline
  \texttt{UOB-REPS} & $\tilde{O}(\sqrt{S^2 A H^4 K})$ & Adv. & \Checkmark & \Checkmark & \XSolidBrush & IS & CI & Optimization \\ 
\hline
 \rowcolor{lightgray} \texttt{POMD} & $\tilde{O}(\sqrt{S^2 A H^4} K^{2/3})$ & Adv. & \Checkmark & \Checkmark & \Checkmark & IS & CI & Closed form\\ 
\hline
\end{tabular}
\end{center}
\label{table: comparison}
\end{table*}

\section{Preliminaries}
\paragraph{Stochastic MDPs.} 
A finite horizon stochastic Markov Decision Process (MDP) $\M$ is defined by a tuple $(\S, \A,H, \{p_h\}_{h=1}^H , \{c_h\}_{h=1}^H)$, where $\S$ and $\A$ are finite state and action spaces with cardinality $S$ and $A$, respectively, and $H\in \mathbb{N}$ is the horizon of the MDP.
On time step $h$, and state $s$, the agent performs an action $a$, transitions to the next state $s'$ according to a time-dependent transition function $p_h(s'\mid s,a)$, and suffers a random cost $C_h(s,a)\in[0,1]$ drawn i.i.d from a distribution with expectation $c_h(s,a)$. 


A stochastic policy $\pi: \mathcal{S}\times[H]\rightarrow \Delta_A$ is a mapping from states and time-step indices to a distribution over actions, i.e., $\Delta_A = \brc*{\pi\in \mathbb{R}^A: \sum_a \pi(a)=1 ,\pi(a)\geq 0}$. The performance of a policy $\pi$ when starting from state $s$ at time $h$ is measured by its value function, which is defined as
\begin{align}
    V_h^\pi(s) = \E\brs*{\sum_{h'=h}^H c_{h'}\br*{s_{h'},a_{h'}}\mid s_h=s,\pi,p}, \label{eq: def value function}
\end{align}
where the expectation is with respect to the randomness of the transition function, the cost function and the policy. 
The $Q$-function of a policy given the state action pair $(s,a)$ at time-step $h$ is defined by
\begin{align}
    Q_h^{\pi}(s,a) = \E \brs*{ \sum_{h'=h}^H c_{h'} (s_{h'},a_{h'}) \mid s_h = s, a_h = a, \pi, p}. \label{eq: def q function}
\end{align}
The two satisfy the following relation:
\begin{align}
    &Q_h^{\pi}(s,a) = c_h(s,a) + p_h(\cdot \mid s,a) V_{h+1}^{\pi},\nonumber\\
    &V^{\pi}_h(s) = \inner{Q^{\pi}_h(s,\cdot),\pi_h(\cdot\mid s)}, \label{eq: relation value-q function}
\end{align}
with $p_h(\cdot \!\mid \! s,a) V \!=\! \sum_{s'}p_h(s'\! \mid\! s,a) V(s')$ for ${V\in\mathbb{R}^S}$, and $\inner*{\cdot,\cdot}$ is the dot product.

An optimal policy $\pi^*$ minimizes the value for all states $s$ and time-steps $h$ simultaneously \cite{puterman2014markov}, and its corresponding optimal value is denoted by $V_h^*(s) = \min_{\pi} V_h^\pi(s),\;$ for all $h\in [H]$. We consider an agent that repeatedly interacts with an MDP in a sequence of $K$ episodes such that the starting state at the $k$-th episode, $s_1^k$, is initialized by a fixed state $s_1$\footnote{for simplicity we fix the initial state, but the results hold when it is drawn from a fixed distribution.}. The agent does not have access to the model, and the costs are received by bandit feedback, i.e., the agent only observes the costs of encountered state-action pairs. At the beginning of the $k$-th episode, the agent chooses a policy $\pi_k$ and samples a trajectory $\brc*{s_h^k,a_h^k,C_h^k(s_h^k,a_h^k)}_{h=1}^H$ by interacting with the stochastic MDP using this policy, where $(s_h^k,a_h^k)$ are the state and action at the $h$-th time-step of the $k$-th episode. The performance of the agent for stochastic MDPs is measured by its \textit{regret} relatively to the value of the optimal policy, defined as 
$\Regret(K')= \sum_{k=1}^{K'} V_1^{\pi_k}(s_1^k) - V_1^*(s_1^k)$ 
for all $K'\in [K]$, and $\pi_k$ is the policy of the agent at the $k$-th episode. 
\paragraph{Adversarial MDPs.} Unlike stochastic MDP, in adversarial MDP, we let the cost to be determined by an adversary at the beginning of every episode, whereas the transition function is fixed. Thus, we denote the MDP at the $k$-th episode by $\M^k = (\S, \A,H, \{p_h\}_{h=1}^H , \{c^k_h\}_{h=1}^H)$. As in~\eqref{eq: def value function},~\eqref{eq: def q function}, we define the value function and $Q$-function of a policy $\pi$ at the $k$-th episode by 
\vspace{-0.3cm}
\begin{align*}
    V_h^{k,\pi}(s) 
    & = 
    \E\brs*{\sum_{h'=h}^H c^k_{h'}\br*{s_{h'},a_{h'}}\mid s_h=s, \pi,p},
    \\
    Q_h^{k,\pi}(s,a) 
    & = 
    \E\brs*{\sum_{h'=h}^H c^k_{h'}\br*{s_{h'},a_{h'}}\mid s_h=s, a_h=a ,\pi,p}.
\end{align*}
Notably, $V_h^{k,\pi}$ and $Q_h^{k,\pi}$ satisfy the relations in relation~\eqref{eq: relation value-q function}. 

We consider an agent which repeatedly interacts with an adversarial MDP in a sequence of $K$ episodes. Each episode starts from a fixed initial state, $s_1^k=s_1$. As in the stochastic case, at the beginning of the $k$-th episode, the agent chooses a policy $\pi_k$ and samples a trajectory $\brc*{s_h^k,a_h^k,c^k_h(s_h^k,a_h^k)}_{h=1}^H$ by interacting with the adversarial MDP. In this case, the performance of the agent is measured by its \textit{regret} relatively to the value of the best policy in hindsight. The objective is to minimize $\Regret(K')= \max_\pi \sum_{k=1}^{K'} V_1^{k,\pi_k}(s_1) - V_1^{k,\pi}(s_1)$ for all $K'\in [K]$. 

\paragraph{Notations and Definitions.} The filtration $\mathcal{F}_k$ includes all events (states, actions, and costs) until the end of the $k$-th episode, including the initial state of the $k+1$ episode. 
We denote by $n_h^k(s,a)$, the number of times that the agent has visited state-action pair $(s,a)$ at the $h$-th step, and by $\bar{X}_k$, the empirical average of a random variable $X$. Both quantities are based on experience gathered until the end of the $k^{th}$ episode and are $\F_k$ measurable. We also define the probability to visit the state-action pair $(s,a)$ at the $k$-th episode at time-step $h$ by $w_{h}^k(s,a)=\Pr\br*{s_h^k=s,a_h^k=a \mid s_1^k,\pi_k,p}$. Since $\pi_k$ is $\filt$ measurable, so is $w_{h}^k(s,a)$. In what follows, we refer to $w_h^k(s,a)$ as the \emph{state-action occupancy measure}.
Furthermore, we define the state visitation frequency of a policy $\pi$ in state $s$ given a transition model $p$ as $d_h^\pi(s;p) = \E \brs*{\ind \brc*{s_h = s} \mid s_1,\pi,p}$. By the two definitions, it holds that $w_h^k(s,a) = d_h^{\pi_k}(s; p ) \pi_h^k(a\mid s)$.

We use $\tilde O(X)$ to refer to a quantity that depends on $X$ up to a poly-log expression of a quantity at most polynomial in $S,A,K,H$ and $\delta^{-1}$.
Similarly, $\lesssim$ represents $\leq$ up to numerical constans or poly-log factors. We define $X\vee Y\triangleq \max\brc*{X,Y}$.

\paragraph{Mirror Descent.}
The mirror descent (MD) algorithm ~\cite{beck2003mirror} is a proximal convex optimization method that minimizes a linear approximation of the objective together with a proximity term, defined in terms of a Bregman divergence between the old and new solution estimates. In our analysis we choose the Bregman divergence to be the Kullback–Leibler (KL) divergence, $d_{KL}$. If $\brc*{f_k}_{k=1}^K$ is a sequence of convex functions $f_k:\R^d \rightarrow \R$, and $C$ is a constraints set, the $k$-th iterate of MD is the following:
\begin{align*}
 &x_{k+1}\in \argmin_{x \in C} \{ t_K\inner*{\nabla f_k(x_k), x - x_k} +  \dkl{x}{x_k} \},
\end{align*}
where $t_K$ is a stepsize. 
In our case, $C$ is the unit simplex $\Delta$, and thus the optimization problem has a closed-form solution,

$$\forall i\in[d],\ x_{k+1}(i) = \frac{x_k(i)\exp \br*{-t_K \nabla_i f_k(x_k)}}{\sum_j x_k(j)\exp \br*{-t_K \nabla_j f_k(x_k)}} . $$
The MD algorithm ensures $\text{Regret}(K')= \sum_{k=1}^{K'} f(x_k) - \min_x f(x) \in O(\sqrt{K})$ for all $K' \in [K]$.

\section{Related Work}\label{sec: related work}

\paragraph{Approximate Policy Optimization:} A large body of work addresses the convergence properties of policy optimization algorithms from an optimization perspective. 
In \citet{kakade2002approximately}, the authors analyzed the Conservative Policy Iteration (CPI) algorithm, an RL variant of the Frank-Wolfe algorithm \cite{scherrer2014local,vieillard2019connections}, and showed it approximately converges to the global optimal solution. Recently,~\citet{liu2019neural} established the convergence of TRPO when neural networks are being used as the function approximators. Furthermore,~\citet{shani2019adaptive} showed that TRPO \cite{schulman2015trust} is in fact a natural RL adaptation of the MD algorithm, and established convergence guarantees. In \cite{agarwal2019optimality}, the authors obtained convergence results for policy gradient based algorithms. However, all of the aforementioned works rely on the strong assumption of a finite concentrability coefficient, i.e., $\max_{\pi,s,h} d_h^{\pi^*}(s;p)/d_h^{\pi}(s;p)<\infty$ . This assumption bypasses the need to address exploration \cite{kakade2002approximately}, and leads to global guarantees based on the local nature of the policy gradients \cite{scherrer2014local}.

\paragraph{Mirror Descent in Adversarial Reinforcement Learning:} 
%
There are two different methodologies for using MD updates in RL. The first and more practical one, is using MD-like updates directly on the policy.
The second is based on optimizing over the space of state-action occupancy measures, that is, visitation frequencies for state-action pairs. 
An occupancy measure represents a policy implicitly. 
For convenience, previous results for regret minimization using MD approaches are summarized in Table~\ref{table: comparison}.

Following the policy optimization approach, and assuming bandit feedback and known dynamics, \citet{neu2010online} (OMDP-BF) established $\tilde O(K^{2/3}) $ regret for the average reward criteria. Alternatively, by assuming full information on the reward functions, unknown dynamics and further assuming both the reward and transition dynamics are linear in some $d$-dimensional features, \citet{cai2019provably} established $\tilde O (\sqrt{d^3 H^4 K})$ regret for their OPPO algorithm. The tabular case is a specific setting of the latter for $d= S A$. 

Instead of directly optimizing the policy,  \citet{zimin2013online} proposed optimizing over the space of state-action occupancy measures with the Relative Entropy Policy Search (O-REPS) algorithm. 
The O-REPS algorithm implicitly learns a policy by solving an MD optimization problem on the primal linear programming formulation of the MDP \cite{altman1999constrained,neu2017unified}.
Considering full information and unknown transitions, \citet{rosenberg2019full} obtained $\tilde O(\sqrt{S^2 A H^4 K})$ regret for their UC-O-REPS algorithm.
Later, \citet{rosenberg2019bandit} extended the algorithm to bandit feedback and obtained a regret of $\tilde{O}(K^{3/4})$.
Recently, by considering an optimistically biased importance sampling estimator, \citet{jin2019learning} established $\tilde O(\sqrt{S^2 A H^4 K})$ for their UOB-REPS algorithm\footnote{Note that in \citet{jin2019learning}, the regret of UOB-REPS is $\tilde O(\sqrt{S^2 A H^2 K})$. However, this is due to the loop-free assumption. To remove this assumption, one needs to multiply the number of states by a factor of $H$.}. The O-REPS variants' updates constitute solving a convex optimization problem with $H S^2 A$ variables on each episode, instead of the closed form solution updates of the direct policy optimization variants.

\paragraph{Value-based Regret Minimization in Episodic RL:}  As opposed to Policy-based methods, there is an extensive literature about regret minimization in episodic MDPs using value-based methods.
The works of \cite{azar2017minimax,dann2017unifying,jin2018q,zanette2019tighter,efroni2019tight} use the optimism in face of uncertainty principle to achieve near-optimal regret bounds.
\citet{jin2018q} also establish a lower bound of $\Omega (\sqrt{S A H^3 K})$.

\section{Mirror Descent for MDPs}

\begin{algorithm}
\caption{POMD with Known Model}\label{alg: uniform TRPO}
\begin{algorithmic}
\REQUIRE $t_K$, $\pi_1$ is the uniform policy.
\FOR{$k=1,..,K$}
    \STATE {\color{gray} \# Policy Evaluation}
    \FOR{ $\forall h=H, H-1,..,1$}
    \FOR{ $\forall s,a\in\sset\times \aset$}
            \STATE $Q^{\pi_k}_h(s,a) = c_h(s,a) + p_h(\cdot \mid s,a) V^{\pi_k}_{h+1}$ 
    \ENDFOR
    \ENDFOR
    \STATE {\color{gray} \# Policy Improvement}
    \FOR{ $\forall s,a,h\in\sset\times \aset \times [H]$}
    \STATE $\pi_h^{k+1}(a|s) \! = \frac{\pi_h^k(a\mid s) \exp\br*{- t_K Q^{\pi_k}_h(s,a)}} {\sum_{a'} \pi_h^k(a'\mid s) \exp\br*{- t_K Q^{\pi_k}_h(s,a')}} $
    \ENDFOR
    
\ENDFOR
\end{algorithmic}
\end{algorithm}

The empirical success of TRPO~\cite{schulman2015trust} and SAC~\cite{haarnoja2018soft} had motivated recent study of MD-like update rules for solving MDPs~\cite{geist2019theory} when the model of the environment is known. Although not explicitly discussed in~\citep{geist2019theory}, such an algorithm can also provide guarantees -- by similar proof technique -- for the case where the cost function is adversarially chosen on each episode. 

Policy Optimization by Mirror Descent (POMD) (see Algorithm~\ref{alg: uniform TRPO}) is conceptually similar to the Policy Iteration (PI) algorithm \cite{puterman2014markov}. It alternates between two stages: (i) policy evaluation, and (ii) policy improvement. Furthermore, much alike PI, POMD updates its policy on the entire state space, given the evaluated $Q$-function. However, as oppose to PI, the policy improvement stage is `soft'. Instead of updating according to the greedy policy, the algorithm applies soft update that keeps the next policy `close' to the current one due to the KL-divergence term.  

Similarly to standard analysis of the MD algorithm, \citet{geist2019theory} established $\tilde{O}(\sqrt{K})$ bound on the regret of Algorithm~\ref{alg: uniform TRPO}. In the next sections, we apply the same approach to problems with unknown model and bandit feedback.

\section{Extended Value Difference Lemma}
The analysis of both stochastic and adversarial cases is built upon a central lemma which we now review. The lemma is a variant of \citep{cai2019provably}[Lemma 4.2], which generalizes classical value difference lemmas. Rewriting it in the following form, enables us to establish our results (proof in Appendix~\ref{sec: difference lemmas}). 

\begin{restatable}[Extended Value Difference]{lemma}{lemmaExtendedValueDiff}\label{lemma: extended value difference}
Let $\pi,\pi'$ be two policies, and $\M = (\sset, \aset, \brc*{p_h}_{h=1}^H, \brc*{c_h}_{h=1}^H)$ and $\M' = (\sset, \aset, \brc*{p'_h}_{h=1}^H, \brc*{c'_h}_{h=1}^H)$ be two MDPs.
Let $\hat Q_h^{\pi,\M}(s,a)$ be an approximation of the $Q$-function of policy $\pi$ on the MDP $\M$ for all $h,s,a$, and let  ${\hat V_h^{\pi,\M}(s) = \inner*{\hat Q_h^{\pi,\M}(s,\cdot) ,\pi_h(\cdot\mid s)}}$.
Then,
\begin{align*}
    & \hat V_1^{\pi,\M}(s_1) - V_1^{\pi',\M'}(s_1)=
    \\
    &\!
    \sum_{h=1}^H \! \E \brs*{ \inner*{\hat Q_h^{\pi,\M}(s_h,\cdot), \pi_h(\cdot \mid s_h) - \pi'_h(\cdot 
    \mid s_h)} \mid s_1,\pi',p'}+
    \\
    &   
    \!\sum_{h=1}^H \!\E \brs*{\hat Q_h^{\pi\!,\M}\!(s_h,\!a_h) \! -\! c_h' \!(s_h,a_h)\! -\! p'_h\!(\cdot | s_h,a_h) \hat V_{h+1}^{\pi\!,\M}\!\mid \! s_1,\!\pi',\! p'}
\end{align*}
where $V_1^{\pi',\M'}$ is the value function of $\pi'$ in the MDP $\M'$.
\end{restatable}

This lemma generalizes existing value difference lemmas. For example, in~\citep{kearns2002near,dann2017unifying} the term ${V_1^{\pi,\M}(s) - V_1^{\pi,\M'}(s)}$ is analyzed, whereas in~\cite{kakade2002approximately} the term ${V_1^{\pi,\M}(s) - V_1^{\pi',\M}(s)}$ is analyzed. In next sections, we will demonstrate how Lemma~\ref{lemma: extended value difference} results in a simple analysis of the POMD algorithm. Importantly, the resulting regret bounds do not depend on concentrability coefficients (see Section~\ref{sec: related work}) nor on any other structural assumptions.

\section{Policy Optimization in Stochastic MDPs}\label{sec: stochastic PO}


We are now ready to analyze the optimistic version of  POMD for stochastic environments~(see Algorithm~\ref{alg: stochastic optimistic TRPO}). Instead of using the known model as in POMD, in Algorithm~\ref{alg: stochastic optimistic TRPO} we use the empirical model to estimate the $Q$-function of an empirical optimistic MDP, with the empirical transition function $\bar p$ and an optimistic cost function $\hat c$.
The empirical transition function $\bar p$ and empirical cost function $\bar c$ are computed by averaging the observed transitions and costs, respectively, that is,
\begin{align*}
    \bar p_h^k(s' \mid s,a)
    & =
    \frac{\sum_{k'=1}^k \ind \br*{s_h^{k'} = s,a_h^{k'} = a,s_{h+1}^{k'} = s'}}{\sum_{k'=1}^k \ind \br*{s_h^{k'} = s,a_h^{k'} = a}\vee 1}
    \\
    \bar c_h^k(s,a)
    & =
    \frac{\sum_{k'=1}^k C_h^{k'}(s,a) \ind \br*{s_h^{k'} = s,a_h^{k'} = a}}{\sum_{k'=1}^k \ind \br*{s_h^{k'} = s,a_h^{k'} = a}\vee 1},
\end{align*}
for every $s,a,s',h,k$.

\begin{algorithm}
\caption{Optimistic POMD for Stochastic MDPs}\label{alg: stochastic optimistic TRPO}
\begin{algorithmic}
\REQUIRE $t_K$, $\pi_1$ is the uniform policy.
\FOR{$k=1,...,K$}
    \STATE Rollout a trajectory by acting $\pi_k$
    \STATE {\color{gray} \# Policy Evaluation}
    \STATE $\forall s \in \sset,\  V^{k}_{H+1}(s) = 0$
    \FOR{ $\forall h = H,..,1$}
        \FOR{$\forall s,a \in  \sset\times \aset $}
            \STATE $\hat c_h^{k-1}(s,a) = \bar c_h^{k-1}(s,a)  - b_h^{k-1}(s,a)$, Eq.~\eqref{eq: bonus term for stochastic}
            \STATE ${Q^{k}_h(s,a) \!=\!  \hat c_h^{k-1}(s,a) \! + \! \bar p_h^{k-1}(\cdot |s,a)V^{k}_{h+1}}$
             \STATE $Q^{k}_h(s,a) = \max\brc*{Q^{k}_h(s,a),0}$
        \ENDFOR
        \FOR{$\forall s \in  \sset$}
            \STATE $V^{k}_{h}(s) = \inner{Q^{k}_h(s,\cdot),\pi_h^k(\cdot \mid s)}$
        \ENDFOR
    \ENDFOR       
    \STATE {\color{gray} \# Policy Improvement}
        \FOR{$\forall h,s,a \in  [H] \times \sset\times \aset$}
        \STATE $\pi_h^{k+1}(a|s) \! = \frac{\pi_h^k(a\mid s) \exp\br*{- t_K Q^{k}_h(s,a)}} {\sum_{a'} \pi_h^k(a'\mid s) \exp\br*{- t_K Q^{k}_h(s,a')}} $
    \ENDFOR
    \STATE  Update counters and empirical model, $n_k,\bar c^k,\bar p^k$
\ENDFOR
\end{algorithmic}
\end{algorithm}

The optimistic cost function $\hat c$ is obtained by adding a bonus term which drives the algorithm to explore, i.e., $\hat c_h^{k-1}(s,a) = \bar c_h^{k-1}(s,a) - b_h^{k-1}(s,a)$, 
and we set 
\begin{align}
    b_h^{k-1}(s,a) 
    =
    b_h^{c,k-1}(s,a) + b_h^{p, k-1}(s,a). \label{eq: bonus term for stochastic}
\end{align} 
The two bonus terms compensate on the lack of knowledge of the true costs and transition model, and are
\begin{align}\label{eq: bonus term choice}
    &b_h^{c,k-1}(s,a) = \tilde{O}\br*{\frac{1}{\sqrt{n_h^{k-1}(s,a)}}},\nonumber\\ 
    &b_h^{p,k-1}(s,a) = \tilde{O}\br*{\frac{\sqrt{S}}{\sqrt{n_h^{k-1}(s,a)}}}.
\end{align}

The following theorem bounds the regret of Algorithm \ref{alg: stochastic optimistic TRPO}. A full proof is found in Appendix~\ref{sec: regret analysis stochastic}.

\begin{restatable}{theorem}{theoremStochasticRegret}\label{theorem: stochastic regret}
For any $K'\in [K]$, setting $t_K=\tilde{O} \br*{H^{-1} K^{-1/2}}$  the regret of Algorithm~\ref{alg: stochastic optimistic TRPO} is bounded by
$$\text{Regret}(K') \leq \Tilde{O}\br*{\sqrt{S^2AH^4 K}}.$$
\end{restatable}

\begin{proofsketch}
We start by decomposing the regret into three terms according to Lemma~\ref{lemma: extended value difference}, and then bound each term separately to get our final regret bound. For any $\pi$,

\begin{align*}
    & \Regret(K')
    = 
    \sum_{k=1}^{K'} V_1^{\pi_k}(s_1^k) - V_1^\pi(s_1^k)
    \\
    & =
    \sum_{k=1}^{K'} V_1^{\pi_k}(s_1^k) - V_1^k(s_1^k) + \sum_{k=1}^{K'} V_1^k(s_1^k) - V_1^\pi(s_1^k)
    \\
    & = 
    \underbrace{\sum_k V_1^{\pi_k}(s_1) - V_1^k(s_1)}_{(\romannumeral 1)} 
    \\
    & + \!\! 
    \underbrace{\sum_{k,h} \E \brs*{ \inner*{Q_h^k(s_h,\cdot), \pi_h^k (\cdot \mid s_h )-\pi_h(\cdot \mid s_h )} \mid s_1,\pi,p}}_{(\romannumeral 2)} 
    \\
    & + \! \! 
    \underbrace{\sum_{k,h} \! \E \big[ Q_h^k(s_h,\!a_h ) \! -\!  c_h \!(s_h,\!a_h)\! -\! p_h\!(\cdot | s_h,\!a_h)  V_{h+1}^{k}\!\mid \! s_1,\!\pi,\! p }_{(\romannumeral 3)} \!\big]
\end{align*}

\paragraph{Term (\romannumeral 1): Bias of $V^k$.} Term (\romannumeral 1) is the bias between the estimated and true value of $\pi_k$, $V^k$ and $V^{\pi_k}$, respectively. Applying Lemma~\ref{lemma: extended value difference}, while using 
$\E\brs*{ X(s_h,a_h)\mid s_1,\pi_k,p} =  \E\brs*{ X(s^k_h,a^k_h)\mid \filt}$ for any $\filt$-measurable function $X \in \mathbb{R}^{S \times A}$,
we bound Term $(\romannumeral 1)$ by
\begin{align*}
    &\sum_{k,h} \! \E\brs*{\Delta c^{k-1}_h(s_h^k,\!a_h^k) \! + \! H \norm{\Delta p^{k-1}_h(\cdot | s_h^k,a_h^k)}_1 \!\mid \! \mathcal{F}_{k-1}}\\
     &+\sum_{k,h} \E\brs*{ b^{c,k-1}_h(s_h^k,a_h^k) + b^{p,k-1}_h(s_h^k,a_h^k) \mid \mathcal{F}_{k-1}}.
\end{align*}

Here $\Delta c^{k-1}_h(s,a) = c_h(s,a) - \bar c^{k-1}_h(s,a)$ and $\Delta p^{k-1}_h(\cdot \mid s,a) = p_h(\cdot \mid s,a) -  \bar p^{k-1}_h(\cdot \mid s,a)$, are the differences between the true cost and transition model to the empirical cost and transition model. Applying Hoeffding's bound and $L_1$ deviation bound~\cite{weissman2003inequalities} we get that w.h.p. for any $s,a$
\begin{align*}
    &\Delta c_h(s,a) \leq \tilde O \br*{\frac{1}{\sqrt{n^{k-1}_h(s,a)}}}= b^r_h(s,a),\\
    &\norm{\Delta p_h(\cdot \mid s,a)}_1\leq \tilde O \br*{\frac{\sqrt{S}}{\sqrt{n^{k-1}_h(s,a)}}}= b^p_h(s,a).
\end{align*}

Thus, w.h.p., we get
\begin{align*}
    (\romannumeral 1) \lesssim 	 \sum_{k=1}^{K'}\sum_{h=1}^H \E\brs*{ \frac{H\sqrt{S}}{\sqrt{n_h^{k-1}(s_h^k,a_h^k)}} \mid \mathcal{F}_{k-1}},
\end{align*}
which can be bounded by $\tilde O\br*{\sqrt{S^2AH^4K}}$ using standard techniques (e.g.,~\citet{dann2017unifying}).


\paragraph{Term (\romannumeral 2): OMD Analysis.} Term (\romannumeral 2) is the linear approximation used in MD optimization procedure. We bound it using an analysis of OMD.
By applying usual OMD analysis (see Lemma~\ref{lemma: OMD orabona}) we have that for any policy $\pi$ and $s,h$,
\begin{align*}
    & \sum_{k=1}^K \inner*{ Q_h^k( \cdot \mid  s), \pi_h^k(\cdot \mid s) - \pi_h(\cdot \mid s) } 
    \\
    & \qquad \leq 
    \frac{\log A}{t_K} + \frac{t_K}{2} \sum_{k=1}^K \sum_a \pi_h^k(a \mid s) (Q_h^k(s,a))^2.
\end{align*}
We plug this back to Term (\romannumeral 2) and use the fact that $0 \le Q_h^k(s,a) \le H$, to obtain
\begin{align*}
    & \text{Term (ii)} =
    \\
    & =
    \sum_{h=1}^H   \E  \brs*{\sum_{k=1}^{K'} \inner*{Q_h^k(s_h,\cdot), \pi_h^k (\cdot | s_h )-\pi_h(\cdot | s_h )} \mid s_1,\pi, p}
    \\
    & \leq \frac{H\log A}{t_K}  +  \frac{t_K H^3 K }{2}.
\end{align*}
By choosing $t_K=\sqrt{2\log A /(H^2 K)}$, we obtain $\text{Term (ii)} \leq \sqrt{2 H^4 K \log A}.$

\paragraph{Term (\romannumeral 3): Optimism.} 
We choose our exploration bonuses in Eq.~\eqref{eq: bonus term choice} such that Term (\romannumeral 3) is non-positive. Specifically, we choose the bonus such that $Q_h^k(s,\!a ) \! -\!  c_h \!(s,\!a)\! -\! p_h\!(\cdot | s,\!a)  V_{h+1}^{k}\leq 0$ for any $s,a$, which implies that $\mathrm{Term (\romannumeral 3)}\leq 0$.
\end{proofsketch}

\begin{remark}
The choice of the bonus term $b_h^{p, k}(s,a)$ is smaller than in~\citep{cai2019provably} by a factor of $\sqrt{S}$. This translates to an improved regret bound by this factor, although~\citep{cai2019provably} assumes full-information feedback on the cost function. 
\end{remark}

\begin{remark}[Bonus vs. Optimistic Model]
Instead of using the additive exploration bonus $b^p$ -- which compensate on the lack of knowledge of transition model -- one can use an optimistic model approach, as in UCRL2~\cite{jaksch2010near}. Following analogous analysis as of Theorem~\ref{theorem: stochastic regret} one can establish the same guarantee $\tilde{O} (\sqrt{S^2AH^4K})$. However, the additive bonus approach results in an algorithm with reduced computational cost.  
\end{remark}

\begin{remark}[Optimism of POMD] Unlike value-based algorithms~(e.g., \citet{jaksch2010near}) $V^k$, the value-function by which POMD improves upon, is not necessarily optimistic relatively to $V^*$. Instead, it is optimistic relatively to the value of $\pi_k$, i.e., $V^k\leq V^{\pi_k}$.
\end{remark}

\section{Policy Optimization in Adversarial MDPs}\label{sec: adveresial po}

\begin{algorithm}
\caption{Optimistic POMD for Adversarial MDPs}\label{alg: adversarial optimistic TRPO}
\begin{algorithmic}
\REQUIRE $t_K$, $\gamma$, $\pi_1$ is the uniform policy.
\FOR{$k=1,...,K$}
    \STATE Rollout a trajectory by acting $\pi_k$
    \FORALL {$h,s$}
        \STATE Compute $u_h^k(s)$ by $\pi_k,\P^{k-1}$, Eq.~\eqref{eq: u upper bound definition}
    \ENDFOR
    \STATE {\color{gray} \# Policy Evaluation}
    \STATE $\forall s \in \sset,\  V^{k}_{H+1}(s) = 0$
    \FOR{ $\forall h = H,..,1$}
        \FOR{$\forall s,a \in  \sset\times \aset $}
            \STATE $\hat c_h^{k}(s,a) = \frac{c^k_h(s,a)\ind\brc*{s=s_h^k,a=a_h^k}}{u^k_h(s)\pi_h^k(a\mid s) +\gamma}$
            \STATE $\hat p^k_h(\cdot |s,a) \in \underset{\hat p_h(\cdot |s,a)\in \P^{k-1}_h(s,a)}{\arg\min}\hat p_h(\cdot |s,a)V^{k}_{h+1}$ 
            \STATE ${Q^{k}_h(s,a) =  \hat c_h^{k}(s,a)  +  \hat p^k_h(\cdot |s,a)V^{k}_{h+1}}$
        \ENDFOR
        \FOR{$\forall s \in  \sset$}
            \STATE $V^{k}_{h}(s) = \inner{Q^{k}_h(s,\cdot),\pi_h^k(\cdot \mid s)}$
        \ENDFOR
    \ENDFOR       
    \STATE {\color{gray} \# Policy Improvement}
    \FOR{$\forall h,s,a \in  [H] \times \sset\times \aset$}
        \STATE $\pi_h^{k+1}(a|s) \! = \frac{\pi_h^k(a\mid s) \exp\br*{- t_K Q^{k}_h(s,a)}} {\sum_{a'}\pi_h^k(a'\mid s) \exp\br*{- t_K Q^{k}_h(s,a')}} $
    \ENDFOR
    \STATE  Update counters and model, $n_k, \bar p^k$
\ENDFOR
\end{algorithmic}
\end{algorithm}

In this section, we turn to analyze an optimistic version of POMD for adversarial environments~(Algorithm~\ref{alg: adversarial optimistic TRPO}). Similarly to the stochastic case, Algorithm~\ref{alg: adversarial optimistic TRPO} follows the POMD scheme, and alternates between policy evaluation, and, soft policy improvement, based on MD-like updates. 

Unlike POMD for stochastic environments, the policy evaluation stage of Algorithm~\ref{alg: adversarial optimistic TRPO} uses different estimates of the instantaneous cost and model. The instantaneous cost is evaluated by a biased importance-sampling estimator, originally suggested by~\citep{neu2015explore}, and recently generalized to adversarial RL settings by~\citep{jin2019learning}, 
\begin{align}
    &\hat c_h^{k}(s,a) = \frac{c^k_h(s,a)\ind\brc*{s=s_h^k,a=a_h^k}}{u^k_h(s)\pi_h^k(a\mid s) +\gamma}, \nonumber\\
    &\text{where}\ u^k_h(s) = \max_{ \hat p\in \mathcal{P}^{k-1}} d^{\pi_k}_h(s;\hat p). \label{eq: u upper bound definition}
\end{align}

Here $\mathcal{P}^{k-1}$ is the set of transition functions obtained by using confidence intervals around the empirical model (see Appendix~\ref{sec:conf-int-p-adv}).

In Algorithm 3 of \citet{jin2019learning}, the authors suggest a computationally efficient dynamic programming based approach for calculating $u^k_h(s)$ for all $h,s$.
The motivation for such an estimate lies in the EXP3 algorithm~\cite{auer2002nonstochastic} for adversarial bandits, which uses an unbiased importance-sampling estimator ${\hat c(a) = \frac{c^k(a) \ind \brc*{a = a^k}}{\pi^k(a)}}$.
Later, \citet{neu2015explore} showed that an optimistically biased estimator ${\hat c(a) = \frac{c^k(a) \ind \brc*{a = a^k}}{\pi^k(a) + \gamma}}$ that motivates exploration can also be used in this setting.
Generalizing the latter estimator to the adversarial RL setting requires to use the estimator $\hat c_h^{k}(s,a) = \frac{c^k_h(s,a)\ind\brc*{s=s_h^k,a=a_h^k}}{d^{\pi_k}_h(s;p)\pi_h^k(a\mid s) +\gamma}$.
However, since the model is unknown, \citet{jin2019learning} uses $u_h^k(s)$ as an upper bound on $d^{\pi_k}_h(s;p)$ which further drives exploration.

Instead of using the empirical model and subtracting a bonus term, Algorithm~\ref{alg: adversarial optimistic TRPO} uses an optimistic model~\cite{jaksch2010near} for the policy evaluation stage. The model by which $Q^k$ is evaluated is the one which results in the smallest loss among possible models,
\begin{align*}
    \hat p^k_h(\cdot |s,a) \in \underset{\hat p_h(\cdot |s,a)\in \P^{k-1}_h(s,a)}{\arg\min}\hat p_h(\cdot |s,a)V^{k}_{h+1}. 
\end{align*}
The solution to this optimization problem can be computed efficiently (see, e.g., \citet{jaksch2010near}).

\begin{remark}[Optimistic Model vs. Additive Exploration Bonus]
Replacing the optimistic model with additive bonuses, we were able to establish $\tilde{O}(K^{3/4})$ regret bound.
It is not clear whether this approach can attain a $\tilde{O}(K^{2/3})$ regret bound, as achieved when using an optimistic model.
\end{remark}

The following theorem bounds the regret of Algorithm \ref{alg: adversarial optimistic TRPO}. A full proof is found in Appendix~\ref{sec: regret analysis adversarial}.
\begin{restatable}{theorem}{theoremAdversarialRegret}\label{theorem: adversarial regret}
For any $K'\in [K]$, setting $\gamma = \tilde{O}(A^{-1/2} K^{-1/3})$ and $t_K = \tilde{O}(H^{-1} K^{-2/3})$, the regret of Algorithm~\ref{alg: adversarial optimistic TRPO} is bounded by
 $$\text{Regret}(K') \leq \tilde O\br*{ H^2S\sqrt{ A}( K^{2/3} + SAK^{1/3})}.$$
\end{restatable}

Central to the analysis are the following claims, formally established in Appendix~\ref{supp: adversarial MDPs}. The first is proved in~\citep{jin2019learning}[Lemma 11], based upon~\citep{neu2015explore}[Lemma 1].

\begin{claim}[\citet{jin2019learning}, Lemma 11]\label{claim: cost diff jin}
Let $\alpha^1,..,\alpha^{K'}$ be a sequence of $\mathcal{F}_{k-1}$ measurable functions such that $\alpha^k\in [0,2\gamma]^{S\times A}$. Then, for any $h$ and $K'\in [K]$, with high probability, ${\sum_{k=1}^{K'} \sum_{s,a} \alpha^k(s,a)\br*{\hat c^{k}_h(s,a)- c^{k}_h(s,a)} \leq \tilde{O}\br*{1}.}$
\end{claim}

\begin{claim}\label{claim: claim v diff}
Let $\alpha^1,..,\alpha^{K'}$ be a sequence of $\mathcal{F}_{k-1}$ measurable functions such that $\alpha^k\in [0,2\gamma]$. For any $s,h$ and $K'\in [K]$, with high probability, 
${\sum_{k=1}^{K'} \alpha^k\br*{V^k_h(s)- V^{\pi_k}_h(s)}\leq \tilde{O}\br*{H}.}$
\end{claim}


Claim~\ref{claim: claim v diff} (see Lemma~\ref{lemma: Weighted Bias of adversarial Q} in the appendix) allows us to derive improved upper bound on $\sum_{k=1}^{K'} V^k_h(s)$ which is crucial to derive the $\tilde O(K^{2/3})$ regret bound. Naively, we can bound $V^k_h(s)$ by recalling it is a value function of an MDP with costs bounded by $1/\gamma$. This leads to the naive bound
\begin{align}
    \sum_{k=1}^{K'} V^k_h(s) \leq K'H/\gamma  \label{eq: without claim 2}.
\end{align}
However, a tighter upper bound can be obtained by applying Claim~\ref{claim: claim v diff} with $\alpha^k = 2\gamma$ for all $k\in [K']$. We have that
\begin{align}
    \sum_{k=1}^{K'} V^k_h(s) \leq\sum_{k=1}^{K'} V^{\pi_k}_h(s)+\frac{H}{\gamma}\leq HK' +\frac{H}{\gamma}, \label{eq: improved by claim 2}
\end{align}
where in the last relation we used the fact that for any $s,h$, $V^{\pi_k}_h(s)\leq H$. In the following proof sketch we apply the later upper bound and demonstrate its importance.

\begin{proofsketch}
We decompose the regret as in Theorem~\ref{theorem: stochastic regret} to  (\romannumeral 1) Bias term,  (\romannumeral 2) OMD term, and (\romannumeral 3) Optimism term. We bound both the Bias and Optimism terms in the appendix while relying on both Claim~\ref{claim: cost diff jin} and Claim~\ref{claim: claim v diff}.

\paragraph{Term (\romannumeral 2): OMD Analysis.} Similarly to the stochastic case, we utilize the usual OMD analysis (Lemma~\ref{lemma: OMD orabona}), which ensures that for any policy $\pi$ and $s,h$,

\begin{align*}
    &\sum_{k=1}^{K'} \inner*{ Q_h^k( \cdot \mid  s), \pi_h^k(\cdot \mid s) - \pi_h(\cdot \mid s) } 
    \\
    & \qquad \leq 
    \frac{\log A}{t_K} + \frac{t_K}{2} \sum_{k=1}^{K'} \sum_a \pi_h^k(a \mid s) (Q_h^k(s,a))^2\\
    &\qquad \leq 
    \frac{\log A}{t_K} + \frac{t_K H}{2\gamma} \sum_{k=1}^{K'} \underbrace{\sum_a \pi_h^k(a \mid s) Q_h^k(s,a)}_{=V_h^k(s)}\\
    &\qquad \leq 
    \frac{\log A}{t_K} + \frac{t_K H}{2\gamma} (HK' +\frac{H}{\gamma}),
\end{align*}
where the second relation holds since $0\leq Q_h^k(s,a)\leq \frac{H}{\gamma}$, and the third relation holds by applying Eq.~\eqref{eq: improved by claim 2}. Plugging this in Term (\romannumeral 2) we get
\begin{align*}
    & \text{Term (ii)} =
    \\
    & =
    \sum_{h=1}^H   \E  \brs*{\sum_{k=1}^{K'} \inner*{Q_h^k(s_h,\cdot), \pi_h^k (\cdot | s_h )-\pi_h(\cdot | s_h )} \mid s_1,\pi, p}
    \\
    & \leq  \frac{H\log A}{t_K} + \frac{t_K H^2}{2\gamma} (HK' +\frac{H}{\gamma}).
\end{align*}

\end{proofsketch}


\section{Discussion}

\paragraph{On-policy vs. Off-policy.} There are two prevalent approaches for policy optimization in practice, on-policy and off-policy. On-policy algorithms, e.g., TRPO~\cite{schulman2015trust}, update the policy based on data gathered following the current policy. This results in updating the policy only in observed states. 
However, in terms of theoretical guarantees, the convergence analysis of this approach requires the strong assumption of finite concentrability coefficient \cite{kakade2002approximately,scherrer2014local,agarwal2019optimality,liu2019neural,shani2019adaptive}.  The assumption arises from the need to acquire global guarantees from the local nature of policy gradients.

The approach taken in this work, is fundamentally different than such on-policy approaches. In each episode, instead of updating the policy only at visited states, the policy is updated over the entire state space, by using all the historical data (in the form of the empirical model). Thus, the analyzed approach bears resemblance to off-policy algorithms, e.g., SAC~\cite{haarnoja2018soft}. There, the authors i) estimate the $Q$-function of the current policy by sampling from a buffer, which contains historical data, and ii) apply an MD-like policy update to states sampled from the buffer. 


The uniform updates of policy-based methods analyzed in this work are in stark contrast to value-based algorithms, such as in \cite{jin2018q,efroni2019tight}, where only observed states are updated. It remains an important open question, whether such updates could also be implemented in a provable policy based algorithm. In the case of stochastic POMD, this may be achieved by using optimistic $Q$-function estimates, instead of estimating the model with UCB-bonus, similarly to \cite{jin2019learning}. There, the authors keep the estimates optimistic with respect to the optimal $Q$-function, $Q^*$. However, in approximate policy optimization, the policy improvement is done with respect to $Q^{\pi_k}$, as described in Algorithm~\ref{alg: uniform TRPO}. Therefore, differently than in \cite{jin2019learning}, such off-policy version would require learning an optimistic $Q^{\pi_k}$ estimator, instead of $Q^*$. 

\paragraph{Policy vs. State-Action Occupancy Optimization.}
In our work, we proposed algorithms which directly optimize the policy. In this scenario, the policy is updated independently at each time step $h$ and state $s$. That is, an optimization problem is solved over the action space in each $h,s$. Therefore, this method requires solving $HS$ optimization problems of size $A$, where each has \emph{a closed form solution} in the tabular setting.

Alternatively, algorithms based on the O-REPS framework \cite{zimin2013online}, follow a different approach and optimize over the state-action occupancy measures instead of directly on policies. In the case of unknown transition model, taking such an approach requires solving a constrained convex optimization problem, later relaxed to a convex optimization problem with only non-negativity constraints~\cite{rosenberg2019full} of size $HS^2A$, in each episode. Unlike the policy optimization approach, this optimization problem \emph{does not have a closed form solution}. Thus, the computational cost of optimizing over the state-action occupancy measures is much worse than the policy optimization one. 

Another significant shortcoming in applying the O-REPS framework is the difficulty to scale it to the function approximation setting. Specifically, in case the state-action occupancy measure is represented by a non-linear function, it is unclear how to solve the constrained optimization problem as defined in~\cite{rosenberg2019full}. Differently than the O-REPS framework, the policy optimization approach scales naturally to the function approximation setting, e.g.,~\citet{haarnoja2018soft}. In this important aspect, policy optimization is preferable.


Interestingly, our work establishes $\tilde O ( \sqrt{K})$ regret when using POMD for the stochastic case, suggesting that policy-based methods are sufficient for solving stochastic MDPs, and thus preferable, compared to the O-REPS framework, as they also enjoy better computational properties. However, in the adversarial case, \citet{jin2019learning} recently established $\tilde O ( \sqrt{K} )$ regret for the UOB-REPS algorithm, where the adversarial variant of POMD only obtains $\tilde O\br*{K^{2/3}}$ regret. Hence, it is of importance to understand whether it is possible to bridge this gap between policy and occupancy measure based methods, or alternatively to show that this gap is in fact a true drawback of policy optimization methods in the adversarial case.

\section{Acknowledgments}
We thank the anonymous reviewers for providing us with very helpful comments.

\bibliography{Bibliography}
\bibliographystyle{icml2020}

\onecolumn

\listofappendices

\begin{appendices}
\section{Additional Notation}\label{sec: additional notation}
We denote, $\bar c$ and $\bar p$, the empirical estimators for $c,p$ respectively. In the adversarial case, we denote $\hat c$ as the importance sampling estimator for the costs and $\hat p$ as the optimistic model. When referring to the estimated MDP, we always denote $\hat \M$, regardless of the estimation method. When using the notation $Q_h^{\pi,p, c}$ and $V_h^{\pi,p,c}$, for some policy $\pi$, transition model $p$ and costs $c$, we refer to the expected Q-function and value function at the $h$-th step, of following the policy $\pi$ on the MDP defined by the transitions $p$ and costs $c$.

\section{Stochastic MDPs}

First, we restate here Algorithm~\ref{alg: stochastic optimistic TRPO} for readability:
\setcounter{algorithm}{1}
\begin{algorithm*}
\caption{Optimistic POMD for Stochastic MDPs}
\begin{algorithmic}
\REQUIRE $t_K$, $\pi_1$ is the uniform policy.
\FOR{$k=1,...,K$}
    \STATE Rollout a trajectory by acting $\pi_k$
    \STATE {\color{gray} \# Policy Evaluation}
    \STATE $\forall s \in \sset,\  V^{k}_{H+1}(s) = 0$
    \FOR{ $\forall h = H,..,1$}
        \FOR{$\forall s,a \in  \sset\times \aset $}
            \STATE $\hat c_h^{k-1}(s,a) = \bar c_h^{k-1}(s,a)  - b_h^{k-1}(s,a)$, Eq.~\eqref{eq: bonus term for stochastic}
            \STATE ${Q^{k}_h(s,a) \!=\!  \hat c_h^{k-1}(s,a) \! + \! \bar p_h^{k-1}(\cdot |s,a)V^{k}_{h+1}}$
             \STATE $Q^{k}_h(s,a) = \max\brc*{Q^{k}_h(s,a),0}$
        \ENDFOR
        \FOR{$\forall s \in  \sset$}
            \STATE $V^{k}_{h}(s) = \inner{Q^{k}_h(s,\cdot),\pi_h^k(\cdot \mid s)}$
        \ENDFOR
    \ENDFOR       
    \STATE {\color{gray} \# Policy Improvement}
        \FOR{$\forall h,s,a \in  [H] \times \sset\times \aset$}
        \STATE $\pi_h^{k+1}(a|s) \! = \frac{\pi_h^k(a\mid s) \exp\br*{- t_K Q^{k}_h(s,a)}} {\sum_{a'} \pi_h^k(a'\mid s) \exp\br*{- t_K Q^{k}_h(s,a')}} $
    \ENDFOR
    \STATE  Update counters and empirical model, $n_k,\bar c^k,\bar p^k$
\ENDFOR
\end{algorithmic}
\end{algorithm*}

In the stochastic case, we use the empirical model:
$$\bar c_h^k(s,a) = \frac{\sum_{k'=1}^k \I\brc*{s_h^{k'} = s,a_h^{k'} = a} c_h^{k'}(s,a)}{n_h^{k'}(s,a) \vee 1}$$
$$\bar p_h^k(s'\mid s,a) = \frac{\sum_{k=1}^{k'} \I\brc*{s_h^{k'} = s,a_h^{k'} = a,s_{h+1}^{k'}=s'}}{ \sum_{k=1}^{k'} \I\brc*{s_h^{k'} = s,a_h^{k'} = a} \vee 1},$$
where $n_h^{k}(s,a) \equiv \sum_{k'=1}^k\I\brc*{s_h^{k'} = s,a_h^{k'} = a}$.

The bonus term in Algorithm~\ref{alg: stochastic optimistic TRPO} is made of a bonus term dedicated to the uncertainty in the rewards and a second term dedicated to the uncertainty in the transition model (see \eqref{eq: bonus term for stochastic}),
\begin{align*}
    b_h^{k-1}(s,a) 
    =
    b_h^{c,k-1}(s,a) + b_h^{p, k-1}(s,a).
\end{align*}

We choose the additive bonus terms as follows (this choice is guided by the need to keep the term in Lemma~\ref{lemma: term 3 stochastic} negative):
\begin{align*}
    &b_h^{k,c}(s,a) = \sqrt{\frac{2\ln \frac{2SAHT}{\delta'}}{n_h^{k-1}(s,a)\vee 1}}\\
    &b_h^{k,pv}(s,a) = H\sqrt{\frac{4S\ln\frac{3SAHT}{\delta'}}{n_h^{k-1}(s,a)\vee 1
    }}.
\end{align*}

\begin{remark}[Bounded Q and value estimators]\label{remark: Q bounded stochastic}
For any $k,h,s,a$, $Q_h^k(s,a)\in[0,H]$ and $V_h^k(s)\in[0,H]$. To see that, first note that by the update rule, we have that for any $k,h,s,a$, $Q_h^k(s,a)\geq 0$. Moreover, using negative bonuses, $Q_h^k$ is always smaller than $Q_h^{\pi_k,\bar p, \bar c}$. Therefore, it is always upper bounded by $H$.
\end{remark}

In the next section, \ref{sec: failure events stochastic}, we deal with all the failure events that can happen while running algorithm~\ref{alg: stochastic optimistic TRPO}, and show that they happen with small probability. Then, in section~\ref{sec: regret analysis stochastic}, we prove Theorem~\ref{theorem: stochastic regret} which establishes the convergence of Algorithm~\ref{alg: stochastic optimistic TRPO}.

\subsection{Failure Events}\label{sec: failure events stochastic}

Define the following failure events.
\begin{align*}
    &F^c_k=\brc*{\exists s,a,h:\ |c_h(s,a) - \bar{c}_h^k(s,a)| \geq \sqrt{\frac{2\ln \frac{2SAHT}{\delta'}}{n_h^{k-1}(s,a)\vee 1}} }\\
    &F^p_k=\brc*{\exists s,a,h:\ \norm{p_h(\cdot\mid s,a)- \bar{p}_h^k(\cdot\mid s,a)}_1 \geq \sqrt{\frac{4S\ln\frac{3SAHT}{\delta'}}{n_h^{k-1}(s,a)\vee 1
    }}}\\
    &F^N_k = \brc*{\exists s,a,h: n_h^{k-1}(s,a) \le \frac{1}{2} \sum_{j<k} w_j(s,a,h)-H\ln\frac{SAH}{\delta'}}.
\end{align*}

Furthermore, the following relations hold.

\begin{itemize}
    \item Let $F^c=\bigcup_{k=1}^K F^c_k.$ Then $\Pr\brc*{F^c}\leq \delta'$, by Hoeffding's inequality, and using a union bound argument on all $s,a$, and all possible values of $n_{k}(s,a)$ and $k$. Furthermore, for $n(s,a)=0$ the bound holds trivially since $C\in[0,1]$. 
    \item Let $F^P=\bigcup_{k=1}^K F^{p}_k.$ Then $\Pr\brc*{ F^p}\leq \delta'$, holds by \citep{weissman2003inequalities} while applying union bound on all $s,a$, and all possible values of $n_k(s,a)$ and $k$. Furthermore, for $n(s,a)=0$ the bound holds trivially. 
    \item Let $F^N=\bigcup_{k=1}^K F^N_k.$ Then, $\Pr\brc*{F^N}\leq \delta'$. The proof is given in \citep{dann2017unifying} Corollary E.4.
\end{itemize}

\begin{lemma}[Good event of the stochastic case]\label{lemma: ucrl failure events}
Setting $\delta'=\frac{\delta}{3}$ then $\Pr\brc{F^c \bigcup F^p\bigcup F^N}\leq \delta$. When the failure events does not hold we say the algorithm is outside the failure event, or inside the good event $G$.
\end{lemma}

\subsection{Regret Analysis - Proof of Theorem~\ref{theorem: stochastic regret}}\label{sec: regret analysis stochastic}

By conditioning our analysis on the good event which was formalized in the previous section (see Lemma~\ref{lemma: ucrl failure events}), we are ready to prove the following theorem, which establishes the convergence of Algorithm~\ref{alg: stochastic optimistic TRPO}.

\theoremStochasticRegret*

\begin{proof}
First, we decompose the regret in the following way,
\begin{align*}
    & \sum_{k=1}^K V_1^{\pi_k}(s_1) - V_1^\pi(s_1)  = \sum_{k=1}^K {V_1^{\pi_k}(s_1) - V_1^k(s_1)} + {V_1^k(s_1) -  V_1^{\pi}(s_1)} \\
    & = \underbrace{\sum_{k=1}^K V_1^{\pi_k}(s_1) - V_1^k(s_1)}_{(\romannumeral 1)} \\
    & + \underbrace{\sum_{k=1}^K \sum_{h=1}^H \E \brs*{ \inner*{Q_h^k(s_h,\cdot), \pi_h^k (\cdot \mid s_h )-\pi_h(\cdot \mid s_h )} \mid s_1 = s, \pi,P}}_{(\romannumeral 2)} \nonumber\\
    & + \underbrace{\sum_{k=1}^K \sum_{h=1}^H \E   \brs*{Q_h^k(s_h,a_h) - c_h(s_h,a_h) - p_h(\cdot \mid s_h,a_h) V_{h+1}^{k} \mid s_1 = s,\pi, P} }_{(\romannumeral 3)},
\end{align*}
where the second relation holds by using the extended value difference lemma (Lemma~\ref{lemma: extended value difference}).

By applying Lemmas~\ref{lemma: term 1 stochastic}, \ref{lemma: term 2 stochastic} and \ref{lemma: term 3 stochastic} to bound each of the above three terms, respectively, we get that conditioned on the good event, for any $K'\in[K]$ and any $\pi$

\begin{align*}
    \sum_{k=1}^{K'} V_1^{\pi_k}(s_1) - V_1^\pi(s_1) \leq \tilde O(\sqrt{S^2AH^4 K}) + \sqrt{2 H^4 K \log A} + 0 \leq  \tilde O(\sqrt{S^2AH^4 K})
\end{align*}

\end{proof}

In what follows we will analyze the each of the three terms separately: Term $(\romannumeral 1)$ is a bias term between the value of the current policy and the estimation of that value, which we bound in Lemma~\ref{lemma: term 1 stochastic}. Term $(\romannumeral 2)$ is the linear approximation term used in the OMD optimization problem. This term will be bounded by the OMD analysis (see Lemma~\ref{lemma: term 2 stochastic}). Term $(\romannumeral 3)$ is an optimism term. It represents the error of our $Q$-function estimation w.r.t. to the $Q$-function obtained by having the real model, and thus, applying the true 1-step Bellman operator. By the optimistic nature of our estimators, this term is negative given the good event (see Lemma~\ref{lemma: term 3 stochastic}).

\begin{lemma}[Bias Term of the Stochastic Case]\label{lemma: term 1 stochastic}
Conditioned on the good event, we have that
\begin{align*}
    \text{Term (i)} = \sum_{k=1}^K V_1^{\pi_k}(s_1) - V_1^k(s_1) \leq O(\sqrt{S^2AH^3T}).
\end{align*}
\end{lemma}
\begin{proof}
By the extended value diffrence lemma~(Lemma \ref{lemma: extended value difference}), we get
\begin{align}
    \sum_{k=1}^K & V_1^{\pi_k}(s_1) - V_1^k(s_1) \nonumber \\
    & = \sum_{k=1}^K \sum_{h=1}^H \E \brs*{ c_h(s_h,a_h) + p_h(\cdot \mid s_h,a_h) V_{h+1}^k - Q_{h+1}^k(s_h,a_h) \mid s_1=s, \pi_k, \M} \nonumber \\ 
    & = \sum_{k=1}^K \sum_{h=1}^H \E \brs*{ c_h(s_h,a_h)  + p_h(\cdot\mid s_h,a_h) V_{h+1}^k \mid s_1=s, \pi_k, \M}  \nonumber \\
    & - \sum_{k=1}^K\sum_{h=1}^H \E \brs*{ \max\brc*{\bar c_h^k(s_h,a_h) - b_h^{k,c}(s_h,a_h) + \bar p_h^{k-1}(\cdot \mid s_h,a_h)V_{h+1}^k - b_h^{k,pv}(s_h,a_h) ,0} \mid s_1=s, \pi_k, \M},\label{eq: term 1 relation 1 stochastic}
\end{align}
where the second relation follows from the update rule of $Q_{h+1}^k$.

First, observe that for any $(k,h,s,a)$
\begin{align}
    c_h&(s,a) + p_h(\cdot \mid s,a)V_{h+1}^k - \max \brc*{\bar c_h^k(s,a) - b_h^{k,c}(s,a) + \bar p_h^{k-1}(\cdot \mid s,a) V_{h+1}^k - b_h^{k,pv}(s,a),0 } \nonumber \\
    &= c_h(s,a) + p_h(\cdot \mid s,a)V_{h+1}^k + \min \brc*{-\bar c_h^k(s,a) + b_h^{k,c}(s,a) - \bar p_h^{k-1}(\cdot \mid s,a) V_{h+1}^k + b_h^{k,pv}(s,a) ,0}\nonumber \\
    &\leq  c_h(s,a) - \bar c_h^k(s,a) + b_h^{k,c}(s,a) + p_h(\cdot \mid s,a)V_{h+1}^k - \bar p_h^{k-1}(\cdot \mid s,a) V_{h+1}^k + b_h^{k,pv}(s,a), \label{eq: term 1 cost and model stochastic}
\end{align} 
where the second relation is by the definition of minimum between two terms.

Conditioning on the good event, we have that for any $(h,k,s,a)$ 
\begin{align}
    c_h(s,a) -\bar c_h^k(s,a) + b_h^{k,c}(s,a) \leq   2b_h^{k,c}(s,a) \label{eq: term 1 cost bound stochastic},
\end{align}
and 
\begin{align}
    p_h&(\cdot\mid s,a)V_{h+1}^k -\bar p_h^{k-1}(\cdot \mid s,a) V_{h+1}^k + b_h^{k,pv}(s,a) \nonumber \\
    & = 
    \br*{p_h(\cdot\mid s,a) -\bar p_h^{k-1}(\cdot \mid s_h,a_h)} V_{h+1}^k + b_h^{k,pv}(s,a)
    \nonumber 
    \\
    & \leq 
    \norm{p_h(\cdot \mid s,a) -\bar p_h^{k-1}(\cdot \mid s,a)}_1 \norm{V_{h+1}^k}_\infty  + b_h^{k,pv}(s,a)
    \nonumber
    \\
    & \leq 
    H \norm{p_h(\cdot \mid s,a) -\bar p_h^{k-1}(\cdot \mid s,a)}_1 + b_h^{k,pv}(s,a) \nonumber\\
    & \leq 
    2b_h^{p}(s,a).
    \label{eq: term 1 model bound stochastic}
\end{align}
See that the second relation is by the Cauchy-Schwartz inequality. The third is by the fact that for any $k,h,s$, $0 \leq V_h^k(s)\leq H$. the last relation holds conditioned on the good event.

Plugging~\eqref{eq: term 1 cost bound stochastic},~\eqref{eq: term 1 model bound stochastic} into~\eqref{eq: term 1 cost and model stochastic} and then back to \eqref{eq: term 1 relation 1 stochastic} we get
\begin{align*}
    & \eqref{eq: term 1 relation 1 stochastic} \leq \sum_{k=1}^K \sum_{h=1}^H \E \brs*{ 2b_h^{k,c}(s_h,a_h) + 2b_h^{k,pv}(s_h,a_h) \mid s_1=s, \pi_k, \M}\\
    & = C\sqrt{\ln \frac{2SAHT}{\delta'}} \sum_{k=1}^K \sum_{h=1}^H \E \brs*{ \sqrt{\frac{1}{n_h^{k-1}(s,a)\vee 1}}+ H\sqrt{\frac{S}{n_h^{k-1}(s,a)\vee 1
    }} \mid s_1=s, \pi_k, \M}\\
    & \leq  CH\sqrt{S}\sqrt{\ln \frac{2SAHT}{\delta'}} \sum_{k=1}^K \sum_{h=1}^H \E \brs*{ \sqrt{\frac{1}{n_h^{k-1}(s,a)\vee 1}} \mid s_1=s, \pi_k, \M}\\
    & =  CH\sqrt{S}\sqrt{\ln \frac{2SAHT}{\delta'}} \sum_{k=1}^K \sum_{h=1}^H \E \brs*{ \sqrt{\frac{1}{n_h^{k-1}(s,a)\vee 1}} \mid \mathcal{F}_{k-1}},
\end{align*}
where in the fourth relation we used the fact that the expectations are equivalent, since at the $k$-th episode we follow the policy $\pi_k$ in the MDP $\mathcal{M}$.

Applying Lemma~\ref{lemma: supp 1 factor and lograthimic factors} we get
\begin{align*}
    \text{Term (i)}\leq \tilde O(\sqrt{S^2AH^4 K}) .
\end{align*}
\end{proof}

\begin{lemma}[OMD Term of the Stochastic Case]\label{lemma: term 2 stochastic}
For any $\pi$
\begin{align*}
    \text{Term (ii)} = \sum_{k=1}^K \sum_{h=1}^H \E \brs*{ \inner*{Q_h^k(s_h,\cdot), \pi_h^k (\cdot \mid s_h )-\pi_h(\cdot \mid s_h )} \mid s_1 = s, \pi,P} \leq \sqrt{2 H^4 K \log A}.
\end{align*}
\end{lemma}
\begin{proof}
This term accounts for the optimization error, bounded by the OMD analysis. 

By standard analysis of OMD with the KL divergence used as the Bregman distance (see Lemma~\ref{lemma: fundamental inequality of OMD}) we have that for any $h\in[H],s\in \sset$ and for policy $\pi$,
\begin{align*}
    \sum_{k=1}^K \inner*{ Q_h^k( \cdot \mid  s), \pi_h^k(\cdot \mid s) - \pi_h(\cdot \mid s) } \leq \frac{\log A}{t_K} + \frac{t_K}{2} \sum_{k=1}^K \sum_a \pi_h^k(a \mid s) (Q_h^k(s,a))^2 
\end{align*}
where $t_K$ is a fixed step size.

By the fact $0\leq Q_h^k(s,a) \leq H$ (see Remark~\ref{remark: Q bounded stochastic}), we have
\begin{align}
\sum_{k=1}^K  \inner*{ Q_h^{k}(s,\cdot),\pi_h^k(\cdot\mid s) - \pi_h(\cdot\mid s)}  \leq \frac{\log A}{t_K} +  \frac{t_K H^2 K }{2}. \label{eq: label term 2 MD}
\end{align}

Thus, we can bound Term (ii) as follows
\begin{align*}
    &\text{Term (ii)} = \sum_{k=1}^K \sum_{h=1}^H \E \brs*{ \inner*{Q_h^k(s_h,\cdot), \pi_h^k (\cdot \mid s_h )-\pi_h(\cdot \mid s_h )} \mid s_1 = s, \pi,p}\\
    &=\sum_{h=1}^H  \E \brs*{\sum_{k=1}^K \inner*{Q_h^k(s_h,\cdot), \pi_h^k (\cdot \mid s_h )-\pi_h(\cdot \mid s_h )} \mid s_1 = s, \pi,p}\\
    &\leq \sum_{h=1}^H \E \brs*{\frac{\log A}{t_K} +  t_K H^2 K  \mid s_1 = s, \pi} = \frac{H\log A}{t_K} +  \frac{t_K H^3 K }{2}.
\end{align*}

See that the first relation holds as the expectation does not depend on $k$. Thus, by linearity of expectation, we can switch the order of summation and expectation. The second relation holds since~\eqref{eq: label term 2 MD} holds for any $s$.

Finally, by choosing $t_K=\sqrt{2\log A /(H^2 K)}$, we obtain

\begin{align}
    \text{Term (ii)} \leq \sqrt{2 H^4 K \log A}.
\end{align}
\end{proof}

\begin{lemma}[Optimism Term of the Stochastic Case]\label{lemma: term 3 stochastic}
Conditioned on the good event, we have that for any $\pi$
$$ \text{Term (\romannumeral 3)} = \sum_{k=1}^K \sum_{h=1}^H \E   \brs*{Q_h^k(s_h,a_h) - c_h(s_h,a_h) - p_h(\cdot \mid s_h,a_h) V_{h+1}^{k} \mid s_1 = s,\pi, P} \leq 0 .$$
\end{lemma}
\begin{proof}

We have that
\begin{align*}
    \text{Term (iii)} = &\sum_{k=1}^K \sum_{h=1}^H \E  \brs*{Q_h^k(s_h,a_h) -c_h(s_h,a_h) - p_h(\cdot\mid s_h,a_h) V_{h+1}^k\mid s_1 = s,\pi, p}.
\end{align*}

By definition,
$$Q_h^k(s,a) = \max \brc*{0, \bar c_h^k(s,a) - b_h^{k,c}(s,a) + \bar p_h^{k-1}(\cdot \mid s,a) V_{h+1}^k - b_h^{k,pv}(s,a)}.$$
Now, by the fact that for any $a,b$, $\max\brc*{a+b,0}\leq \max\brc*{a,0} + \max\brc*{b,0}$, we have that

$$Q_h^k(s,a) \leq \max \brc*{0, \bar c_h^k(s,a) - b_h^{k,c}(s,a)} + \max\brc*{0,\bar p_h^{k-1}(\cdot \mid s,a) V_{h+1}^k - b_h^{k,pv}(s,a)}.$$

Therefore, for any $k,h,s,a$, 

\begin{align}
    Q_h^k&(s,a) - c_h(s,a) - p_h(\cdot \mid s,a) V_{h+1}^k 
     \nonumber\\
    & \leq 
    \max \brc*{0, \bar c_h^k(s,a) - b_h^{k,c}(s,a)} + \max\brc*{0,\bar p_h^{k-1}(\cdot \mid s,a) V_{h+1}^k - b_h^{k,pv}(s,a)} - c_h(s,a) -  p_h(\cdot\mid s,a) V_{h+1}^k
         \nonumber\\
    & = 
    \max \brc*{- c_h(s,a), \bar c_h^k(s,a)-   c_h(s,a) - b_h^{k,c}(s,a)}  
         \nonumber\\
    &  + 
    \max\brc*{- p_h(\cdot\mid s,a) V_{h+1}^k,\br*{\bar p_h^{k-1}(\cdot \mid s,a)- p_h(\cdot\mid s,a)} V_{h+1}^k - b_h^{k,pv}(s,a)} 
         \nonumber\\
    & \leq 
    \max \brc*{0, \bar c_h^k(s,a)-   c_h(s,a) - b_h^{k,c}(s,a)}  
         \nonumber\\
    &  + 
    \max\brc*{0,\br*{\bar p_h^{k-1}(\cdot \mid s,a)- p_h(\cdot\mid s,a)} V_{h+1}^{k-1} - b_h^{k,pv}(s,a)} \label{eq: term 3 stochastic 1}
\end{align}

Conditioned on the good event, we have that for any $(k,h,s,a)$,
\begin{align}
    \bar  c_h(s_h,a_h) -c_h(s_h,a_h) -b_h^{k,c}(s,a) \leq 0. \label{eq: term 3 1.1}
\end{align}

Furthermore,
\begin{align}
    &\br*{\hat p_h^{k-1}(\cdot \mid s_h,a_h) -  p_h(\cdot\mid s_h,a_h)} V_{h+1}^{k} -b_h^{k,pv}(s_h,a_h) \nonumber\\
    &\leq \norm{\hat p_h^{k-1}(\cdot \mid s_h,a_h) -  p_h(\cdot\mid s_h,a_h)}_1\norm{ V_{h+1}^{k}}_\infty -b_h^{k,pv}(s_h,a_h) \nonumber\\
    &\leq H\norm{\hat p_h^{k-1}(\cdot \mid s_h,a_h) -  p_h(\cdot\mid s_h,a_h)}_1 -b_h^{k,pv}(s_h,a_h)\leq 0.\label{eq: term 3 1.2}
\end{align}
The first relation holds by Holder's inequality. The second relation holds by the updating rule, which keeps $ 0\leq V_{h+1}^{\pi_k,\hat P,\hat c}\leq H$ (see Remark~\ref{remark: Q bounded stochastic}). The third relation holds conditioning on the good event. 

Plugging~\eqref{eq: term 3 1.1},~\eqref{eq: term 3 1.2} into~\eqref{eq: term 3 stochastic 1} we get
$$ \text{Term (\romannumeral 3)}\leq 0 .$$

\end{proof}
\newpage

\section{Adversarial MDPs}\label{supp: adversarial MDPs}

First, we restate here Algorithm~\ref{alg: adversarial optimistic TRPO} for readability:
\begin{algorithm}
\caption{Optimistic POMD for Adversarial MDPs}
\begin{algorithmic}
\REQUIRE $t_K$, $\gamma$, $\pi_1$ is the uniform policy.
\FOR{$k=1,...,K$}
    \STATE Rollout a trajectory by acting $\pi_k$
    \FORALL {$h,s$}
        \STATE Compute $u_h^k(s)$ by $\pi_k,\P^{k-1}$, Eq.~\eqref{eq: u upper bound definition}
    \ENDFOR
    \STATE {\color{gray} \# Policy Evaluation}
    \STATE $\forall s \in \sset,\  V^{k}_{H+1}(s) = 0$
    \FOR{ $\forall h = H,..,1$}
        \FOR{$\forall s,a \in  \sset\times \aset $}
            \STATE $\hat c_h^{k}(s,a) = \frac{c^k_h(s,a)\ind\brc*{s=s_h^k,a=a_h^k}}{u^k_h(s)\pi_h^k(a\mid s) +\gamma}$
            \STATE $\hat p^k_h(\cdot |s,a) \in \underset{\hat p_h(\cdot |s,a)\in \P^{k-1}_h(s,a)}{\arg\min}\hat p_h(\cdot |s,a)V^{k}_{h+1}$ 
            \STATE ${Q^{k}_h(s,a) =  \hat c_h^{k}(s,a)  +  \hat p^k_h(\cdot |s,a)V^{k}_{h+1}}$
        \ENDFOR
        \FOR{$\forall s \in  \sset$}
            \STATE $V^{k}_{h}(s) = \inner{Q^{k}_h(s,\cdot),\pi_h^k(\cdot \mid s)}$
        \ENDFOR
    \ENDFOR       
    \STATE {\color{gray} \# Policy Improvement}
    \FOR{$\forall h,s,a \in  [H] \times \sset\times \aset$}
        \STATE $\pi_h^{k+1}(a|s) \! = \frac{\pi_h^k(a\mid s) \exp\br*{- t_K Q^{k}_h(s,a)}} {\sum_{a'}\pi_h^k(a'\mid s) \exp\br*{- t_K Q^{k}_h(s,a')}} $
    \ENDFOR
    \STATE  Update counters and model, $n_k, \bar p^k$
\ENDFOR
\end{algorithmic}
\end{algorithm}

We define the costs of the online MDP at the $k$-th episode, for each $h\in[H], s \in \sset, a \in \aset$, and for any $\pi_h^k$
\begin{align}\label{eq: costs estimator online}
\hat c_h^k(s,a) \triangleq \frac{c_h^k(s,a)\I\br*{s_h^k=s,a_h^k=a}}{u_h^k(s)\pi_h^k(a \mid s)+\gamma}
\end{align}

We also define the following optimistic model, $\hat p_h^k(\cdot \mid s,a)$, which is the solution to the following optimization problem:
\begin{align*}
   \hat p^k_h(\cdot |s,a) \in \underset{\hat p_h(\cdot |s,a)\in \P^{k-1}_h(s,a)}{\arg\min}\hat p_h(\cdot |s,a)V^{k}_{h+1},
\end{align*}
where $\P_h^k(s,a)$ is defined in \eqref{eq: probability set definition} Finally, as for the stochastic case, we denote the empirical estimator of the transition function as
$$\bar p_h^k(s'\mid s,a) = \frac{\sum_{k=1}^{k'} \I\brc*{s_h^{k'} = s,a_h^{k'} = a,s_{h+1}^{k'}=s'}}{\sum_{s''} \sum_{k=1}^{k'} \I\brc*{s_h^{k'} = s,a_h^{k'} = a,s_{h+1}^{k'}=s''}}.$$

\begin{remark}[Bounded Q and value estimators]\label{remark: Q bounded adversarial}
For any $k,h,s,a$, $Q_h^k(s,a)\in[0,H/\gamma]$ and $V_h^k(s)\in[0,H/\gamma]$. To see that, first note that $\hat c_h^k(s,a) \in [0,1/\gamma]$. By the fact that the estimators for the Q-function and value function are always calculated w.r.t. to some transition model $\hat p$, we get that the estimators are bounded as suggested.
\end{remark}

The following lemmas, Lemma~\ref{lemma: Bias of adversarial costs} and  Lemma~\ref{lemma: Weighted Bias of adversarial Q}, will be essential to establish regret bounds for Algorithm~\ref{alg: adversarial optimistic TRPO}. In the main body of the paper, we refer to these lemmas as claim~\ref{claim: cost diff jin} and claim~\ref{claim: claim v diff}, respectively. Lemma~\ref{lemma: Bias of adversarial costs} is a very close adaptation of \citep{jin2019learning}[Lemma 11], which in itself based on \citep{neu2015explore}[Lemma 1]. Lemma~\ref{lemma: Weighted Bias of adversarial Q} relies upon applying Lemma~\ref{lemma: Bias of adversarial costs}.

\begin{restatable}[Bias of the adversarial costs,~\citealt{jin2019learning}]{lemma}{lemmaBiasAdversarialCosts}\label{lemma: Bias of adversarial costs}
Let $\alpha^1,...,\alpha^K$ be a sequence of functions, such that $\alpha^k \in [0,2\gamma]^{S\times A}$ is $\filt$-measurable for all $k$. Let $u_h^k(s)>0$ for any $k,h,s$. Then,
With probability of at least $1 - \delta$, for any $h \in [H]$,
\begin{align*}
    \sum_{k=1}^K \sum_{s,a} 
    \alpha^k(s,a) \br*{\hat c_h^k(s,a) 
    - 
    \frac{d_h^k(s)}{u_h^k(s)} c_h^k(s,a)} 
    \le 
    \ln \frac{1}{\delta}.
\end{align*}
\end{restatable}

The full proof of Lemma~\ref{lemma: Bias of adversarial costs} is given in section \ref{sec: useful lemmas}.

\begin{restatable}[Bias of the adversarial value functions]{lemma}{lemmaBiasAdversarialQV}\label{lemma: Weighted Bias of adversarial Q} Let $\alpha^1,...,\alpha^K$ be a sequence of functions, such that $\alpha^k \in [0,1]$ is $\filt$-measurable for all $k$. Furthermore, assume that for all  $k,h,s$  $u_h^k(s)> d_h^k(s)\geq 0$. Then, with probability of at least $1-\delta$, for any fixed $h\in [H]$ and $s\in\sset$,
\begin{align*}
    2\gamma \sum_{k=1}^{K} \alpha^k \br*{V_h^{\pi_k,p,\hat c}(s) - V_h^{\pi_k}(s)} 
    \le 
    H \ln \frac{H}{\delta},
\end{align*}
where $V_h^{\pi_k,p,\hat c}$ is the value of following the policy $\pi_k$ at the $h$-th step, on the MDP defined by the transitions $p$ and costs $\hat c$ (as defined in Appendix~\ref{sec: additional notation}).
\end{restatable}


\begin{proof}

For any $(h,s)$ we have
\begin{align}
    \sum_{k=1}^K &   2\gamma  \alpha^k \br*{V_h^{\pi_k,p,\hat c}(s) - V_h^{\pi_k}(s)}
    \nonumber \\
    & =
    \sum_{k=1}^K \sum_{h'=h}^H  2\gamma  \alpha^k \E \brs*{\hat c_{h'}^k(s_{h'},a_{h'}) - c_{h'}^k(s_{h'},a_{h'})\mid s_h=s,\pi_k, p}
     \nonumber\\
    & \le
    \sum_{k=1}^K  \sum_{h'=h}^H  2\gamma \alpha^k\E \brs*{\hat c_{h'}^k(s_{h'},a_{h'}) - \frac{d_h^k(s_{h'})}{u_h^k(s_{h'})} c_{h'}^k(s_{h'},a_{h'})\mid s_h=s, \pi_k, p} 
    \nonumber \\
    & =
     \sum_{h'=h}^H \sum_{k=1}^K \sum_{s_{h'}} \sum_{a_{h'}}  2\gamma \alpha^k \Pr(s_{h'},a_{h'} \mid s_h=s, \pi_k, p) \br*{\hat c_{h'}^k(s_{h'},a_{h'}) - \frac{d_h^k(s_{h'})}{u_h^k(s_{h'})} c_{h'}^k(s_{h'},a_{h'})}. \label{eq: claim 2 relation 1}
\end{align}
The first relation holds by Corollary \ref{corollary: value difference}, as both value functions are measured w.r.t. the same dynamics and are defined over the same policy. The second relation holds by the fact $c_h^k(s,a)\geq 0 $ and by the fact that by the assumptions of the lemma, for any $h,k,s$,  $d_h^k(s) \leq u_h^k(s)$.

Now, observe that $  \Pr(s_{h'},a_{h'} \mid s_h=s, \pi_k, p)\in [0,1]$ and are measurable functions w.r.t. $\filt$. For any $h'\in \brc*{h,..,H}$ we set $\alpha_{h'}^k(s_{h'},a_{h'}) = 2 \gamma \alpha^k \Pr(s_{h'},a_{h'} \mid s_h=s, \pi_k, p)$ where $\alpha_{h'}^k(s_{h'},a_{h'})\in [0,2\gamma]$. By this definition we have that
\begin{align*}
    \eqref{eq: claim 2 relation 1} = \sum_{h'=h}^H \sum_{k=1}^K \sum_{s_{h'}} \sum_{a_{h'}} \alpha_{h'}^k(s_{h'},a_{h'}) \br*{\hat c_{h'}^k(s_{h'},a_{h'}) - \frac{d_h^k(s_{h'})}{u_h^k(s_{h'})} c_{h'}^k(s_{h'},a_{h'})}.
\end{align*}

For any $h'\in \brc*{h,..,H}$ we apply Lemma~\ref{lemma: Bias of adversarial costs}, take a union bound and bound $H-h\leq H$ to get
\begin{align*}
    \sum_{h'=h}^H \sum_{k=1}^K \sum_{s_{h'}} \sum_{a_{h'}} \alpha_{h'}^k(s_{h'},a_{h'}) \br*{\hat c_{h'}^k(s_{h'},a_{h'}) - \frac{d_h^k(s_{h'})}{u_h^k(s_{h'})} c_{h'}^k(s_{h'},a_{h'})}\leq H\ln{\frac{H}{\gamma}}
\end{align*}
w.p. $1-\delta$.


\end{proof}


\subsection{Failure Events}\label{sec: failure events adversarial}
In this section we define the high probability bounds which are later in use in the proof of Theorem~\ref{theorem: adversarial regret}. We divide the failure event into two different kinds of failure event: basic failure events which are independent on each other, and conditioned failure event which holds conditioned on the basic failure event. 

The next sections are ordered in the following way: we first define the basic failure event and the resulting basic good event. Then, we describe the consequences of this basic good event. Finally, we describe the conditioned failure events, which rely on the consequences of the basic good event. By combining all failure events, we define the global failure event. In the proof, we condition our analysis on the event the global failure event does not hold. We also refer to this event as the good event.

\subsubsection{Basic failure events:}
\begin{align*}
    &F_k^p=\brc*{\exists s,a,s',h:\ \abs{ p_h(s' \mid s,a) - \bar p_h^{k}(s'\mid s,a)}\geq 2\sqrt{\frac{\bar p_h^{k}(s' \mid s,a) (1- \bar p_h^{k}(s' \mid s,a))\ln \br*{\frac{HSAK}{4\delta'}}}{(n_h^{k}(s,a)-1) \vee 1}} + \frac{14\ln \br*{\frac{HSAK}{4\delta'}}}{3 \br*{(n_h^{k}(s,a) - 1) \vee 1}} }\\
    &F^N_k = \brc*{\exists s,a,h: n_h^{k-1}(s,a) \le \frac{1}{2} \sum_{j<k} w_j(s,a,h)-H\ln\frac{SAH}{\delta'}}\\
    &F^{c}_{k'} = \brc*{\exists s,a,h: \sum_{k=1}^{k'} \hat  c^k_h(s,a)-\frac{d_h^k(s)}{u_h^k(s)}c^k_h(s,a) \geq \frac{\ln \frac{SAHK}{\delta'}}{2\gamma}}\\
\end{align*}

Furthermore, the following relations hold.

\begin{itemize}
    \item Let $F^p = \bigcup_{k=1}^K F_k^p$. Then, $\Pr\brc*{ F^p}\leq \delta'$, by (\citealp{maurer2009empirical}, Theorem 4) and union bounds.
    \item Let $F^N=\bigcup_{k=1}^K F^N_k.$ Then, $\Pr\brc*{F^N}\leq \delta'$. The proof is given in \citep{dann2017unifying} Corollary E.4.
     \item Fix  $k'\in[K]$ and let $u_h^k(s)>0$ for all $k\in[k']$. Fix $s,a,h$,  and let $\delta''>0$. For any $k\in[k']$, define $\alpha_h^k(s',a') = 2\gamma \ind\brc*{s'=s,a'=a}$ (which is a constant function, and hence measurable). We have that
     \begin{align*}
         &2\gamma \sum_{k=1}^{k'} \hat c^k_h(s,a)-\frac{d_h^k(s)}{u_h^k(s)} c^k_h(s,a)\\
     &=\sum_{k=1}^{k'} \sum_{s',a'}\alpha_h^k(s',a') (\hat  c^k_h(s',a')-\frac{d_h^k(s)}{u_h^k(s)}c^k_h(s',a')) \leq \ln\frac{H}{\delta''},
     \end{align*}
     by Lemma~\ref{lemma: Bias of adversarial costs} w.p. $1-\delta''$ for any $h$. Taking union bound on $s,a$ and setting $\delta'' = \frac{\delta'}{SAK}$, we get that $\Pr\brc*{F_{k'}^c}\leq \frac{\delta'}{K}$. Finally, let $F^c=\bigcup_{k'=1}^K F_{k'}^c$. By union bound, $\Pr\brc*{F^c}\leq \delta'.$
 \end{itemize}
 
 Finally, setting $\delta'=\frac{\delta}{6}$, and denote $F^{\text{basic}} \triangleq F^p\bigcup F^N \bigcup F^c$. Then, by union bound $\Pr\brc{F^{\text{basic}}}\leq \frac{\delta}{2}$.

\begin{lemma}[Basic good event of the adversarial case]\label{lemma: adversarial basic failure events}
Denote $G^{\text{basic}}\triangleq \neg F^{\text{basic}}$, then $Pr\brc{G^{\text{basic}}}\geq 1-\frac{\delta}{2}$. When $G^{\text{basic}}$ occurs, we say that the basic good event holds.
\end{lemma}

\subsubsection{Consequences Conditioning on the Basic Good event}
\label{sec:conf-int-p-adv}

First,
 for any $k,h,s,a$, we define the set
\begin{align}\label{eq: probability set definition}
        \P_h^k(s,a) =  \brc*{\hat p_h(\cdot \mid s,a) : \forall  s'  \enskip \abs{\hat p_h(s' \mid s,a) - \bar p_h^{k}(s' \mid s,a)}\leq \epsilon_k(s' \mid s,a) , p_h(s' \mid s,a) \geq 0  , \sum_{s'} \hat p_h(s' \mid s,a) = 1} ,
    \end{align}
    where $$ \epsilon_h^k(s'\mid s,a)\triangleq 2\sqrt{\frac{\bar p_h^{k}(s' \mid s,a) (1- \bar p_h^{k}(s' \mid s,a))\ln \br*{\frac{HSAK}{4\delta'}}}{(n_h^{k}(s,a)-1) \vee 1}} + \frac{14\ln \br*{\frac{HSAK}{4\delta'}}}{3 \br*{ (n_h^{k}(s,a) - 1) \vee 1}} $$

By using this definition conditioned on the basic good event, we get the following lemma from \cite{jin2019learning}[Lemma 8],
\begin{lemma}\label{lemma: probability ball around p}
Conditioned on the basic good event, for all $k,h,s,a,s'$ and for all $\hat p_h^k(\cdot \mid s,a) \in \P_h^{k-1}(s,a)$, there exists constants $C_1,C_2>0$ for which we have that 
$$ \abs*{\hat p_h^k(s' \mid s,a) - p_h(s' \mid s,a)} =  C_1\sqrt{\frac{ p_h(s' \mid s,a) \ln \br*{\frac{HSAK}{4\delta'}}}{(n_h^{k}(s,a)-1) \vee 1}} + \frac{C_2\ln \br*{\frac{HSAK}{4\delta'}}}{ (n_h^{k}(s,a)-1) \vee 1}. $$
\end{lemma}

\begin{lemma} \label{lemma: estimated q to q of true dynamics}
Conditioned on the basic good event, for any $(s,a)\in \mathcal{S}\times \mathcal{A}$, $h\in [H], k\in [K]$
\begin{align*}
    V^k_h(s) \leq V_h^{\pi_k,p,\hat c}(s),
\end{align*}
where $V_h^{\pi_k,p,\hat c}$ is defined in Appendix~\ref{sec: additional notation}.
\end{lemma}

\begin{proof}
By definition of the update rule,
\begin{align}
    Q^{k}_h(s,a) 
    = 
    \hat{c}^k_h(s,a) + \hat p_h^k(\cdot \mid s,a) V_{h+1}^k. \label{eq: update rule q adverseraial term 2}
\end{align}

By the description of the algorithm, for each value, we solve the following minimization problem, for any $k,h,s,a$
$$ \hat p_h^k(\cdot \mid s,a) \in \argmin_{ p_h^{k}(\cdot \mid s,a) \in \P_h^{k-1}(s,a)}  p_h^{k}(\cdot \mid s,a) V_h^k . $$ 
Therefore, by conditioning on the good event and by lemma~\ref{lemma: probability ball around p}, for any $k,h,s,a$ the following holds
\begin{align}
&\hat p^k_h (\cdot \mid s,a) V_{h+1}^k \leq  p_h (\cdot \mid s,a) V_{h+1}^k. \label{eq: term 2 optimistic adverserial con1}
\end{align}

By plugging in \eqref{eq: term 2 optimistic adverserial con1} in \eqref{eq: update rule q adverseraial term 2}, we get
\begin{align}\label{eq: term 2 optimistic adversarial con2}
Q^{k}_h(s,a) \leq \hat{c}^k_h(s,a) + p_h (\cdot \mid s,a) V_{h+1}^k.
\end{align}

Now, note that for $h=H$ using the fact that $V_{H+1}^k=0$ for any $k,s$, we obtain,
$$Q^{k}_H(s,a) = \hat{c}^k_H(s,a) + \hat p_h (\cdot \mid s,a) V_{H+1}^k = \hat{c}^k_H(s,a) = Q_H^{\pi_k,p,\hat c}(s,a),$$
and therefore, for any $k,s$ and policy $\pi_k$
$$V_H^k(s) \leq V_H^{\pi_k,p,\hat c}.$$

Using the above inequality, by backward recursion on $h=H,H-1,...,1$ on \eqref{eq: term 2 optimistic adversarial con2}, we get for any $k,h,s,a$
\begin{align*}
     Q_h^k(s,a) & \leq \hat c_h^k(s,a) + p_h(\cdot \mid s,a) V_{h+1}^k \leq \hat c_h^k(s,a) + p_h(\cdot \mid s,a) V_{h+1}^{\pi_k,p,\hat c} = Q_h^{\pi_k,p,\hat c}(s,a),
\end{align*}
where in the second inequality we used the fact that $p_h$, $V_{h+1}^k$ and $V_{h+1}^{\pi_k,p,\hat c}$ are all non-negative.

Furthermore, 
$$ V_h^k(s) \leq V_h^{\pi_k,p,\hat c}(s),$$
follows immediately.
\end{proof}

\subsubsection{Conditioned failure events}\label{sec: conditioned failure event}

\begin{align*}
    &F^{\hat c} = \brc*{\exists h:   \sum_{k'=1}^K  \sum_{h=1}^H \br*{\E \brs*{\sum_s \sum_a d_h^k(s) \pi_k(a\mid s) \hat c_h^k(s,a) \mid \filt} - \sum_s \sum_a d_h^k(s) \pi_k(a\mid s)\hat c_h^k(s,a)} \geq H\sqrt{K\ln \frac{H}{2\delta'}}}\\
    & F_{k'}^{v,MD} = \brc*{\exists h,s: \sum_{k=1}^K  V_h^k(s) - V_h^{\pi_k}(s) \geq \frac{H}{2\gamma} \ln \frac{H^2 SK}{\delta'}}\\
    & F_{k'}^{v,1} = \brc*{\exists s,a,s',h: \sum_{h=1}^H \sum_{s,a,s'}  \sum_{k=1}^K \sqrt{\frac{p_h(s' \mid s,a)}{(n_h^{k-1}(s,a)-1) \vee 1}} \Pr(s_h \! =\!  s,a_h \! =\! a | s_1,\! \pi_k, p) \br*{V_{h+1}^k(s') \!- \! V_{h+1}^{\pi_k}(s')} \geq \frac{H^2 \! S^2\! A}{\gamma} \ln \frac{H^2S^2AK}{\delta'}}\!\!\!\\
    & F_{k'}^{v,2} = \brc*{\exists s,a,s',h: \sum_{h=1}^H \sum_{s,a,s'} \sum_{k=1}^K \frac{\Pr(s_h = s,a_h = a \mid s_1, \pi_k, p)}{ (n_h^{k-1}(s,a)-1) \vee 1}  \br*{V_{h+1}^k(s') - V_{h+1}^{\pi_k}(s')}  \geq \frac{H^2 S^2 A}{2\gamma} \ln \frac{H^2S^2AK}{\delta'}}.
\end{align*}

 \begin{itemize}    
     \item Fix $h$ and let $\delta'>0$. Conditioning on the the basic good event $G^{\text{basic}}$ for any $k,h$,
\begin{align*} 
    \sum_s \sum_a d_h^k(s) \pi_h^k(a\mid s) \hat c_h^k(s,a) 
    & =
    \sum_s \sum_a d_h^k(s) \pi_h^k(a\mid s) \frac{c_h^k(s,a) \I(s_h^k=s,a_h^k=a)}{u_h^k(s) \pi_h^k(a\mid s) + \gamma} 
    \\
    & \leq 
    \sum_s \sum_a \I(s_h^k=s,a_h^k=a)
    = 
    1,
\end{align*}
where we conditioned on the event $ G^{\text{basic}}$ in which $d_h^k(s) \leq u_h^k(s)$, 
Therefore, $\sum_s \sum_a d_h^k(s) \pi_h^k(a\mid s) \hat c_h^k(s,a)\in[0,1]$.
Furthermore, observe
\begin{align*}
    \sum_{k'=1}^K  \sum_{h=1}^H \br*{\E \brs*{\sum_s \sum_a d_h^k(s) \pi_k(a\mid s) \hat c_h^k(s,a) \mid \filt} - \sum_s \sum_a d_h^k(s) \pi_k(a\mid s)\hat c_h^k(s,a)},
\end{align*}
is a martingale-difference sequence. Thus, by Azuma-Hoeffding and taking union bound for all $H$, we have that for any $K$, $Pr\brc*{F^{\hat c} \mid \neg F^{\text{basic}}}\leq \delta''=\frac{\delta'}{H}$.

\item Fix $k'\in[K],h,s$ and let $\delta''>0$. Now, set for any $k\in[k']$, $\alpha^k = 1\in[0,1]$ (constant and thus measurable). Furthermore, conditioned on the basic good event, we have that for any $k,h,s$, $u_h^k(s) > d_h^k(s)\geq 0$. Thus, by applying Lemma~\ref{lemma: Weighted Bias of adversarial Q}, we get that w.p. $1-\delta''$
     \begin{align*}
        \sum_{k=1}^{k'} \alpha^k \br*{V_h^{\pi_k,p,\hat c}(s) - V_h^{\pi_k}(s)} \leq\frac{H}{2\gamma} \ln \frac{H}{\delta''}.
     \end{align*}
     Now, conditioned on the basic good event, by Lemma~\ref{lemma: estimated q to q of true dynamics} we have
     \begin{align*}
        \sum_{k=1}^{k'} \alpha^k \br*{V_h^k(s) - V_h^{\pi_k}(s)} \leq\frac{H}{2\gamma} \ln \frac{H}{\delta''}.
     \end{align*}
    Taking union bounds on $h,s$ and setting $\delta'' = \frac{\delta'}{HSK}$, we get that $\Pr \brc*{F_{k'}^{v,MD}}\leq \frac{\delta'}{K}$. Finally, let $F^{v,MD}=\bigcup_{k'=1}^K F_{k'}^{v,MD}$. By union bound, $\Pr\brc*{F^{v,MD}}\leq \delta'.$
 \item Fix $k'\in[K], s,a,s',h$ and let $\delta''>0$. Now, set for any $k\in[k']$, $$\alpha^k(s') = \sum_{s,a} \sqrt{\frac{p_h(s' \mid s,a)}{(n_h^{k-1}(s,a)-1) \vee 1}} \Pr(s_h = s,a_h=a \mid s_1, \pi_k, p)\in[0,1],$$ and note that it is $\filt$-measurable. Furthermore, conditioned on the basic good event, we have that for any $k,h,s$, $u_h^k(s) > d_h^k(s)\geq 0$. Thus, by applying Lemma~\ref{lemma: Weighted Bias of adversarial Q}, we get that w.p. $1-\delta''$, for any fixed $s'$
     \begin{align*}
        \sum_{k=1}^{k'} \alpha^k(s') \br*{V_h^{\pi_k,p,\hat c}(s') - V_h^{\pi_k}(s')} \leq\frac{H}{2\gamma} \ln \frac{H}{\delta''}.
     \end{align*}
    Now, conditioned on the basic good event, by Lemma~\ref{lemma: estimated q to q of true dynamics} we have
     \begin{align*}
        \sum_{k=1}^{k'} \alpha^k(s') \br*{V_h^k(s') - V_h^{\pi_k}(s')} \leq\frac{H}{2\gamma} \ln \frac{H}{\delta''},
     \end{align*}
     w.p. $1-\delta''$.
     Taking union bound on $s',h$ and setting $\delta'' = \frac{\delta'}{SHK}$, we get that w.p. $1-\frac{\delta'}{K}$
      \begin{align*}
        \sum_{h=1}^H \sum_{s'}\sum_{k=1}^K \alpha^k(s') \br*{V_h^k(s) - V_h^{\pi_k}(s)} \leq\frac{H^2 S}{2\gamma} \ln \frac{H^2 S K }{\delta'},
     \end{align*}
     or in other words, $\Pr\brc*{F_{k'}^{v,1}}\leq \frac{\delta'}{K}$. Finally, let $F^{v,1}=\bigcup_{k'=1}^K F_{k'}^{v,1}$. By union bound, $\Pr\brc*{F^{v,1}}\leq \delta'.$
\item By following the same proof of event $F^{v,1}$ (i.e., by applying Lemma~\ref{lemma: Weighted Bias of adversarial Q}), but using $\alpha^k(s') = \sum_{s,a} \frac{\Pr(s_h=s,a_h=a \mid s_1, \pi_k, p)}{ (n_h^{k-1}(s_h,a_h)-1) \vee 1}\in[0,1]$ for any $s'$, we get that $\Pr\brc*{F^{v,2}}\leq \delta'$.
\end{itemize}

Now, denote the conditioned event, $F^{\text{conditioned}}\triangleq F^{\hat c}\bigcup F^{v,MD} \bigcup F^{v,1} \bigcup F^{v,2}$.

Next, we set $\delta'=\frac{\delta}{8}$. 
Then, by union bound $ \Pr\brc{ F^{\text{conditioned}} \mid \neg F^{\text{basic}}} \leq \frac{\delta}{2}$.

\begin{lemma}[Conditioned good event of the adversarial case]\label{lemma: adversarial conditioned failure events}
Denote $G^{\text{conditioned}}\triangleq \neg F^{\text{conditioned}}$, then $\Pr\brc{ G^{\text{conditioned}} \mid G^{\text{basic}}}\geq 1-\frac{\delta}{2}$.
\end{lemma}

\subsubsection{Global Failure Events}

In this section, we combine both the basic and conditioned failure events into a single global failure event. The global failure event accounts for all failure events which can occur in the adversarial MDP case. Specifically, in our analysis we will always assume that none of the failure events occurs, which happens with probability of at least $1-\delta$ since

$$\Pr\brc{\neg F^{\text{conditioned}} \bigcap \neg F^{\text{basic}}} = \Pr\brc{\neg F^{\text{conditioned}} \mid \neg F^{\text{basic}}}\Pr\brc{\neg F^{\text{basic}}} \geq \br*{1-\frac{\delta}{2}}\br*{1-\frac{\delta}{2}}\geq 1-\delta , $$
where we used the facts that $\Pr\brc{\neg F^{\text{basic}}}\geq 1-\frac{\delta}{2} $ by Lemma~\ref{lemma: adversarial basic failure events}, and $\Pr\brc{\neg F^{\text{conditioned}} \mid \neg F^{\text{basic}}} \geq 1-\frac{\delta}{2}$ by Lemma~\ref{lemma: adversarial conditioned failure events}.

\begin{lemma}[Good event of the adversarial case]\label{lemma: adversarial failure events}
Denote $G\triangleq G^{\text{conditioned}} \bigcap G^{\text{basic}} = \neg F^{\text{conditioned}} \bigcap \neg F^{\text{basic}}$, then $Pr\brc{G}\geq 1-\delta$. When $G$ occurs, we say the algorithm outside the failure event or inside the good event.
\end{lemma}

\subsection{Regret Analysis - Proof of Theorem~\ref{theorem: adversarial regret}}\label{sec: regret analysis adversarial}
By conditioning our analysis on the good event which was formalized in the previous sections (see Lemma~\ref{lemma: adversarial failure events}), we are ready to prove the following theorem, which establishes the convergence of Algorithm~\ref{alg: adversarial optimistic TRPO}.

\theoremAdversarialRegret*

\begin{proof}
First, we decompose the regret in the following way
\begin{align*}
    & \sum_{k=1}^K V_1^{\pi_k}(s_1) - V_1^\pi(s_1)  = \sum_{k=1}^K {V_1^{\pi_k}(s_1) - V_1^k(s_1)} + {V_1^k(s_1) -  V_1^{\pi}(s_1)} \\
    & = \underbrace{\sum_{k=1}^K V_1^{\pi_k}(s_1) - V_1^k(s_1)}_{(\romannumeral 1)} \\
    & + \underbrace{\sum_{k=1}^K \sum_{h=1}^H \E \brs*{ \inner*{Q_h^k(s_h,\cdot), \pi_h^k (\cdot \mid s_h )-\pi_h(\cdot \mid s_h )} \mid s_1 = s, \pi,p}}_{(\romannumeral 2)} \nonumber\\
    & + \underbrace{\sum_{k=1}^K \sum_{h=1}^H \E   \brs*{Q_h^k(s_h,a_h) - c_h(s_h,a_h) - p_h(\cdot \mid s_h,a_h) V_{h+1}^{k} \mid s_1 = s,\pi, p} }_{(\romannumeral 3)},
\end{align*}
where the second relation holds by using the extended value difference lemma (Lemma~\ref{lemma: extended value difference}).

By applying Lemmas~\ref{lemma: term 1 adversarial}, \ref{lemma: term 2 adversarial} and \ref{lemma: term 3 adversarial} to bound each of the above three terms, respectively, we get that conditioned on the good event (see Lemma~\ref{lemma: adversarial failure events}), for any $K'\in[K]$,
\begin{align*}
    \text{Regret}(K) &\leq \tilde O\br*{\sqrt{S^2A H^4 K} + \gamma HSAK + \sqrt{H^2 K} + \frac{H^2 S}{\gamma}} \\
    & + \tilde O\br*{\frac{H\log A}{t_K} + \frac{2 t_K H^3 }{\gamma^2} + \frac{ t_K H^3 K }{\gamma}   + \frac{H}{\gamma}}
\end{align*}
By choosing $t_K = \tilde O\br*{\sqrt{\log A/(H^2)} K^{-2/3}}$ and $\gamma = \tilde O\br*{A^{-1/2}K^{-1/3}}$, we obtain
\begin{align*}
    \text{Regret}(K) &\leq \tilde O\br*{\sqrt{S^2A H^4 K} +  HS\sqrt{A}K^{2/3} + \sqrt{H^2 K} + H^2 S^2 A^{3/2} K^{1/3} } \\
    & + \tilde O\br*{\sqrt{ H^4 \log A } K^{2/3} + \sqrt{A \log A} H^2 + \sqrt{A \log A H^4} K^{2/3} +  H \sqrt{A} K^{1/3}}\\
    & \leq O\br*{ \sqrt{ H^4 S^2 A} K^{2/3} + H^2 S^2 A^{3/2} K^{1/3}},
\end{align*}
which concludes the proof.
\end{proof}

The decomposition in the proof of Theorem~\ref{theorem: adversarial regret} is the same as in the stochastic case. The analysis is different here due the different nature of the estimators for the costs and transition model. Again, term $(\romannumeral 1)$ is a bias term between the value of the current policy and the estimation of that value, which is bounded in Lemma~\ref{lemma: term 1 adversarial}. Term $(\romannumeral 2)$ is the linear approximation term used in the OMD optimization problem. This term will be bounded by the OMD analysis (see Lemma~\ref{lemma: term 2 adversarial}). Term $(\romannumeral 3)$ is an optimism term. It represents the error of our $Q$-function estimation w.r.t. to the $Q$-function obtained by having the real model, and thus, applying the true 1-step Bellman operator. By the optimistic nature of our estimators, this term is (almost) negative given the good event (see Lemma~\ref{lemma: term 3 adversarial}).

\begin{lemma}[Bias Term of the Adversarial Case]\label{lemma: term 1 adversarial}
Conditioned on the good event,
\begin{align*}
    \text{Term (\romannumeral 1)}= \sum_{k=1}^K V_1^{\pi_k}(s_1) - V_1^k(s_1)\leq \tilde O\br*{\sqrt{S^2A H^4 K} + \gamma HSAK + \sqrt{H^2 K} + \frac{H^2 S}{\gamma}}
\end{align*}
\end{lemma}
\begin{proof}
First, by Lemma~\ref{lemma: extended value difference}, the following relations hold,
\begin{align}
    \sum_{k=1}^K & V_1^{\pi_k}(s_1) - V_1^k(s_1) \nonumber \\
    & = \underbrace{\sum_{k=1}^K \sum_{h=1}^H \E \brs*{ c_h^k(s_h,a_h) - \hat c_h^k(s_h,a_h)  \mid s_1=s, \pi_k, P}}_{(A)}  \nonumber \\
    & + \underbrace{\sum_{k=1}^K\sum_{h=1}^H \E \brs*{ p_h(\cdot\mid s_h,a_h) V_{h+1}^k -\hat  p_h^k(\cdot\mid s_h,a_h) V_{h+1}^k \mid s_1=s, \pi_k, P}}_{(B)},\label{eq: term 1 relation 1 adversarial}
\end{align}

\paragraph{Term (A).}
For any $(k,h,s,a)$,
\begin{align*}
c_h^k(s,a) - \hat c_h^k(s,a)  &= c_h^k(s,a) - \E \brs*{\hat c_h^k(s,a) \mid \filt} + \E \brs*{\hat c_h^k(s,a) \mid \filt} - \hat c_h^k(s,a) \\
& = c_h^k(s,a) \br*{ 1 - \frac{d_h^k(s)\pi_h^k(a \mid s)}{u_h^k(s)\pi_h^k(a\mid s) + \gamma  } }  +\E \brs*{\hat c_h^k(s,a) \mid \filt} - \hat c_h^k(s,a) \\
& = c_h^k(s,a) \br*{ \frac{u_h^k(s)\pi_h^k(a\mid s) - d_h^k(s)\pi_h^k(a \mid s) + \gamma}{u_h^k(s)\pi_h^k(a\mid s) + \gamma  } }  + \E \brs*{\hat c_h^k(s,a) \mid \filt} - \hat c_h^k(s,a).
\end{align*}

By plugging back to the first term of \eqref{eq: term 1 relation 1 adversarial},
\begin{align}
    \sum_{k=1}^K & \sum_{h=1}^H \E \brs*{ c_h^k(s_h,a_h) - \hat c_h^k(s_h,a_h)  \mid s_1=s, \pi_k, \M}  \nonumber\\
    & = \sum_{k=1}^K  \sum_{h=1}^H \E \brs*{ c_h^k(s,a) \br*{ \frac{u_h^k(s)\pi_h^k(a\mid s) - d_h^k(s)\pi_h^k(a \mid s) + \gamma}{u_h^k(s)\pi_h^k(a\mid s) + \gamma  } } \mid s_1=s, \pi_k, \M} \nonumber \\
    & + \sum_{k=1}^K  \sum_{h=1}^H \E \brs*{\E \brs*{\hat c_h^k(s,a) \mid \filt} - \hat c_h^k(s,a)  \mid s_1=s, \pi_k, \M}\nonumber \\
    & = \sum_{k=1}^K  \sum_{h=1}^H \sum_s \sum_a d_h^k(s) \pi_h^k(a\mid s)  c_h^k(s,a) \br*{ \frac{u_h^k(s)\pi_h^k(a\mid s) - d_h^k(s)\pi_h^k(a \mid s) + \gamma}{u_h^k(s)\pi_h^k(a\mid s) + \gamma  } } \nonumber \\
    & + \sum_{k=1}^K  \sum_{h=1}^H \sum_s \sum_a d_h^k(s) \pi_k(a\mid s)\br*{\E \brs*{\hat c_h^k(s,a) \mid \filt} - \hat c_h^k(s,a)}. \label{eq: term 1 realtion 2 adversarial}
\end{align}

First, we deal with the first term in \eqref{eq: term 1 realtion 2 adversarial},
\begin{align}
    \sum_{k=1}^K & \sum_{h=1}^H \sum_s \sum_a d_h^k(s) \pi_h^k(a\mid s)  c_h^k(s,a) \br*{ \frac{u_h^k(s)\pi_h^k(a\mid s) - d_h^k(s)\pi_h^k(a \mid s) + \gamma}{u_h^k(s)\pi_h^k(a\mid s) + \gamma  } } \nonumber\\
    & \leq \sum_{k=1}^K  \sum_{h=1}^H \sum_s \sum_a  \br*{u_h^k(s) \pi_h^k (a\mid s) -   d_h^k(s) \pi_h^k (a\mid s)} + \gamma HSAK \nonumber\\
    & = \sum_{k=1}^K  \sum_{h=1}^H \sum_s \br*{u_h^k(s) -   d_h^k(s)} + \gamma HSAK,\label{eq: term 1 relation 3 part 1 bound adversarial}
\end{align}
where in the inequality we use the fact that by conditioning on the good event, for any $k,h,s$, $d_h^k(s)\leq u_h^k(s)$, and therefore for any $k,h,s,a$, 
$ 
\frac{d_h^k(s) \pi_h^k(a\mid s)}{u_h^k(s) \pi_h^k(a\mid s)+\gamma} \leq 1 
$

As for the second term in \eqref{eq: term 1 realtion 2 adversarial}, conditioning on the good event we have that
\begin{align}\label{eq: term 1 relation 3 part 2 bound adversarial}
    \sum_{k=1}^K  \sum_{h=1}^H \sum_s \sum_a d_h^k(s) \pi_k(a\mid s)\br*{\E \brs*{\hat c_h^k(s,a) \mid \filt} - \hat c_h^k(s,a)} \leq H\sqrt{2K \ln \frac{H}{\delta}}.
\end{align}

By combining \eqref{eq: term 1 relation 3 part 1 bound adversarial} and \eqref{eq: term 1 relation 3 part 2 bound adversarial}, we obtain
\begin{align*}
    \sum_{k=1}^K &\sum_{h=1}^H \E \brs*{ c_h^k(s_h,a_h) - \hat c_h^k(s_h,a_h)  \mid s_1=s, \pi_k, \M}
    \\
    & \leq
    \sum_{k=1}^K  \sum_{h=1}^H \sum_s \br*{u_h^k(s) -   d_h^k(s)} + \gamma HSAK +  H\sqrt{2K \ln \frac{H}{\delta}}
    \\
    & \le
    O \br*{ H S \sqrt{A T \ln \frac{SAHK}{\delta'}}} + \gamma HSAK +  H\sqrt{2K \ln \frac{H}{\delta}},
\end{align*}
where the last relation follows from Lemma \ref{lemma: occupancy-measures-distance}.

\paragraph{Term (B).}
Now, its left to address the second term of \eqref{eq: term 1 relation 1 adversarial}.
Consider the following,

\begin{align}\label{eq: term 1B adversarial difference}
    p_h&(\cdot \mid s_h,a_h) V_{h+1}^k -\hat p_h^k(\cdot \mid s_h,a_h) V_{h+1}^k \nonumber \\
    & = \br*{p_h(\cdot \mid s_h,a_h) -\hat p_h^k(\cdot \mid s_h,a_h) }V_{h+1}^k\nonumber  \\
            & \leq \sum_{s'}\br*{C_1\sqrt{\frac{p_h(s' \mid s_h,a_h) \ln \frac{HSAK}{\delta}}{(n_h^{k-1}(s_h,a_h) -1 ) \vee 1}}+ \frac{C_2\ln\frac{HSAK}{\delta}}{ (n_h^{k-1}(s_h,a_h) -1 ) \vee 1}  }V_{h+1}^k(s')\nonumber  \\
    & = \sum_{s'}\br*{C_1\sqrt{\frac{p_h(s' \mid s_h,a_h) \ln \frac{HSAK}{\delta}}{(n_h^{k-1}(s_h,a_h) -1 ) \vee 1}}+ \frac{C_2\ln\frac{HSAK}{\delta}}{ (n_h^{k-1}(s_h,a_h) -1 ) \vee 1} }V_{h+1}^{\pi_k}(s')\nonumber   \\
    & +  \sum_{s'} \br*{C_1\sqrt{\frac{p_h(s' \mid s_h,a_h) \ln \frac{HSAK}{\delta}}{(n_h^{k-1}(s_h,a_h) -1 ) \vee 1}}+ \frac{C_2\ln\frac{HSAK}{\delta}}{ (n_h^{k-1}(s_h,a_h) -1 ) \vee 1} }\br*{V_{h+1}^k(s') - V_{h+1}^{\pi_k}(s')}.
\end{align}
The second transition is by the fact $V_h^k$ is positive and by the conditioning on the good event and applying Lemma~\ref{lemma: probability ball around p}. The third transition is by the fact for any $k,h,s,a$, $n_h^{k-1}(s,a) \leq n_h^{k-1}(s,a)$.

First, we deal with the first term. Conditioning on the good event, we have for any $(k,s,a,h)$ 
\begin{align}
\sum_{s'}&\br*{C_1\sqrt{\frac{p_h(s' \mid s_h,a_h) \ln \frac{HSAK}{\delta}}{(n_h^{k-1}(s_h,a_h) -1 ) \vee 1}}+ \frac{C_2\ln\frac{HSAK}{\delta}}{ (n_h^{k-1}(s_h,a_h) -1 ) \vee 1} }V_{h+1}^{\pi_k}(s') \nonumber \\
& \leq H\sum_{s'}\br*{C_1\sqrt{\frac{p_h(s' \mid s_h,a_h) \ln \frac{HSAK}{\delta}}{(n_h^{k-1}(s_h,a_h) -1 ) \vee 1}}+ \frac{C_2\ln\frac{HSAK}{\delta}}{ (n_h^{k-1}(s_h,a_h) -1 ) \vee 1} } \nonumber \\
& \leq C_1 HS \sqrt{\frac{\sum_{s'}p_h(s' \mid s_h,a_h) \ln \frac{HSAK}{\delta}}{S (n_h^{k-1}(s_h,a_h) -1 ) \vee 1}}+ \frac{C_2HS\ln\frac{HSAK}{\delta}}{ (n_h^{k-1}(s_h,a_h) -1 ) \vee 1} \nonumber \\
& = C_1 H \sqrt{\frac{S \ln \frac{HSAK}{\delta}}{ (n_h^{k-1}(s_h,a_h) -1 ) \vee 1}}+ \frac{C_2HS\ln\frac{HSAK}{\delta}}{ (n_h^{k-1}(s_h,a_h) -1 ) \vee 1}. \label{eq: term iB adv}
\end{align}
In the first transition we used the fact that $V_h^{\pi_k}$ is positive and bounded by $H$ for any $k,h,s'$. The second transition is by Jensen's inequality and the fact that the square root is concave.

By summing as done in~\eqref{eq: term 1 relation 1 adversarial} we get
\begin{align}
   & \eqref{eq: term iB adv}= \sum_{k=1}^K \sum_{h=1}^H \E \brs*{C_1 H \sqrt{\frac{S \ln \frac{HSAK}{\delta}}{ (n_h^{k-1}(s_h,a_h) -1 ) \vee 1}}+ \frac{C_2HS\ln\frac{HSAK}{\delta}}{ (n_h^{k-1}(s_h,a_h) -1 ) \vee 1}  \mid s_1=s, \pi_k, \M} \nonumber\\
    & = C_1 H\sqrt{S}\sqrt{\ln \frac{2SAHK}{\delta'}} \sum_{k=1}^K \sum_{h=1}^H \E \brs*{ \sqrt{\frac{1}{(n_h^{k-1}(s,a)-1)\vee 1
    }} \mid \filt}\nonumber\\
    & +  C_2 H S \ln \frac{2SAHK}{\delta'} \sum_{k=1}^K \sum_{h=1}^H \E \brs*{ \frac{1}{(n_h^{k-1}(s,a)-1)\vee 1
    } \mid \filt} \nonumber\\
    & \leq C_1 H\sqrt{2S}\sqrt{\ln \frac{2SAHK}{\delta'}} \sum_{k=1}^K \sum_{h=1}^H \E \brs*{ \sqrt{\frac{1}{n_h^{k-1}(s,a)\vee 1
    }} \mid \filt}\nonumber\\
    & +  2 C_2 H S \ln \frac{2SAHK}{\delta'} \sum_{k=1}^K \sum_{h=1}^H \E \brs*{ \frac{1}{n_h^{k-1}(s,a)\vee 1
    } \mid \filt }.
\end{align}
Note that in the first relation we used the fact that the expectations are equivalent, since at the $k$-th episode we follow the policy $\pi_k$ in the MDP $\mathcal{M}$. The third relation holds by the fact that for any $n\geq 0 $, it holds that $\frac{1}{(n-1) \vee 1} \leq \frac{2}{n \vee 1}$.

Finally, applying Lemma~\ref{lemma: supp 1 factor and lograthimic factors} and Lemma~\ref{lemma: supp 1 1/N  factor and lograthimic factors} and excluding constant and logarithmic factors in $K$, we get
\begin{align*}
    \sum_{k=1}^K& \sum_{h=1}^H \E \brs*{C_1 H \sqrt{\frac{S \ln \frac{HSAK}{\delta}}{ (n_h^{k-1}(s_h,a_h) -1 ) \vee 1}}+ \frac{C_2 HS\ln\frac{HSAK}{\delta}}{ (n_h^{k-1}(s_h,a_h) -1 ) \vee 1}  \mid s_1=s, \pi_k, \M} \leq \tilde O\br*{\sqrt{S^2AH^4K}}.
\end{align*}

Now, consider the second term of \eqref{eq: term 1B adversarial
difference}.
\begin{align*}
    \sum_{k=1}^K& \sum_{h=1}^H \E \brs*{ \sum_{s'} \br*{C_1 \sqrt{\frac{p_h(s' \mid s_h,a_h) \ln \frac{HSAK}{\delta}}{(n_h^{k-1}(s_h,a_h) -1 ) \vee 1}}+ \frac{C_2\ln\frac{HSAK}{\delta}}{ (n_h^{k-1}(s_h,a_h) -1 ) \vee 1} }\br*{V_{h+1}^k(s') - V_{h+1}^{\pi_k}(s')} \mid s_1=s, \pi_k, \M} \\
    & = \sum_{k=1}^K\sum_{h=1}^H \sum_{s'} \sum_{s_h,a_h} \Pr(s_h,a_h \mid s_1 = s, \pi_k, p)\br*{C_1 \sqrt{\frac{p_h(s' \mid s_h,a_h) \ln \frac{HSAK}{\delta}}{(n_h^{k-1}(s_h,a_h) -1 ) \vee 1}}+ C_2\frac{\ln\frac{HSAK}{\delta}}{(n_h^{k-1}(s_h,a_h) -1 ) \vee 1} }\br*{V_{h+1}^k(s') - V_{h+1}^{\pi_k}(s')}\\
    & = \sum_{h=1}^H \sum_{s_h,a_h} \sum_{s'} \sum_{k=1}^K C_1\sqrt{\frac{p_h(s' \mid s_h,a_h) \ln \frac{HSAK}{\delta}}{(n_h^{k-1}(s_h,a_h) -1 ) \vee 1}} \Pr(s_h,a_h \mid s_1, \pi_k, p) \br*{V_{h+1}^k(s') - V_{h+1}^{\pi_k}(s')}\\
    & + \sum_{h=1}^H \sum_{s_h,a_h} \sum_{s'} \sum_{k=1}^K \frac{C_2 \ln\frac{HSAK}{\delta}}{ (n_h^{k-1}(s_h,a_h) -1 ) \vee 1} \Pr(s_h,a_h \mid s_1, \pi_k, p) \br*{V_{h+1}^k(s') - V_{h+1}^{\pi_k}(s')} \\
    & = C_1 \sqrt{\ln\frac{HSAK}{\delta}} \sum_{h=1}^H \sum_{s_h,a_h} \sum_{s'} \sum_{k=1}^K \sqrt{\frac{p_h(s' \mid s_h,a_h)}{(n_h^{k-1}(s_h,a_h) -1 ) \vee 1}} \Pr(s_h,a_h \mid s_1, \pi_k, p) \br*{V_{h+1}^k(s') - V_{h+1}^{\pi_k}(s')}\\
    & + C_2 \ln\frac{HSAK}{\delta}\sum_{h=1}^H \sum_{s_h,a_h} \sum_{s'} \sum_{k=1}^K \frac{\Pr(s_h,a_h \mid s_1, \pi_k, p)}{ (n_h^{k-1}(s_h,a_h) -1 ) \vee 1}  \br*{V_{h+1}^k(s') - V_{h+1}^{\pi_k}(s')}\\
    & \le C_1 \sqrt{\ln\frac{HSAK}{\delta}} \sum_{h=1}^H  \sum_{s'} \sum_{k=1}^K \br*{\sum_{s_h,a_h} \frac{\Pr(s_h,a_h \mid s_1, \pi_k, p)}{\sqrt{(n_h^{k-1}(s_h,a_h) -1 ) \vee 1}}} \br*{V_{h+1}^k(s') - V_{h+1}^{\pi_k}(s')}\\
    & + C_2 \ln\frac{HSAK}{\delta}\sum_{h=1}^H \sum_{s'} \sum_{k=1}^K \br*{\sum_{s_h,a_h} \frac{\Pr(s_h,a_h \mid s_1, \pi_k, p)}{ (n_h^{k-1}(s_h,a_h) -1 ) \vee 1}}  \br*{V_{h+1}^k(s') - V_{h+1}^{\pi_k}(s')}.
\end{align*}

Next, conditioned on the good event, and specifically on events $F^{v,1},F^{v,2}$ we have that
\begin{align*}
    \sum_{k=1}^K& \sum_{h=1}^H \E \brs*{ \sum_{s'} \br*{C_1 \sqrt{\frac{p_h(s' \mid s_h,a_h) \ln \frac{HSAK}{\delta}}{(n_h^{k-1}(s_h,a_h) -1 ) \vee 1}}+ \frac{C_2\ln\frac{HSAK}{\delta}}{ (n_h^{k-1}(s_h,a_h) -1 ) \vee 1} }\br*{V_{h+1}^k(s') - V_{h+1}^{\pi_k}(s')} \mid s_1=s, \pi_k, \M} \\
    &\leq O\br*{\frac{H^2 S}{\gamma}\ln\frac{HSAK}{\delta}}.
\end{align*}

Finally, by combining the bounds, we get that 

$$ \text{Term B} \leq \tilde O\br*{\sqrt{S^2AH^3T} + \frac{H^2 S}{\gamma}}. $$

The result holds by combining the two above terms.

\end{proof}

\begin{lemma}[OMD Term of the Adversarial Case]\label{lemma: term 2 adversarial}
Conditioned on the good event, for any $pi$,
\begin{align*}
    \text{Term (\romannumeral 2)}= \sum_{k=1}^K  \sum_{h=1}^H \E \brs*{\inner*{ Q_h^k(s_h,\cdot),\pi_h^k(\cdot\mid s_h) - \pi_h(\cdot\mid s_h)} \mid s_1 = s, \pi, p} \leq \frac{H\log A}{t_K} + \frac{2 t_K H^3 }{\gamma^2} + \frac{2 t_K H^3 K }{\gamma} .
\end{align*}
\end{lemma}
This term accounts for the optimization error, bounded by the OMD analysis when the KL-divergence is used as the Bregman divergence. 

By Lemma~\ref{lemma: fundamental inequality of OMD}, we have that for any $h\in[H],s\in \sset$ and for policy $\pi$,
\begin{align}
    \sum_{k=1}^K \inner*{ Q_h^k( \cdot \mid  s), \pi_h^k(\cdot \mid s) - \pi_h(\cdot \mid s) } \leq \frac{\log A}{t_K} + \frac{t_K}{2} \sum_{k=1}^K \sum_a \pi_h^k(a \mid s) (Q_h^k(s,a))^2  \label{eq:term 2 main term }
\end{align}
where $t_K$ is a fixed step size.

Now, conditioning on the good event, the following holds,
\begin{align*}
    ( Q_h^k(s,a))^2 &= \br*{\hat c_h^k(s,a)+ \hat p_h^k (\cdot \mid s,a) V_{h+1}^k}^2\\ 
    & \leq 2\br*{\hat c_h^k(s,a)}^2 + 2\br*{\hat p_h^k (\cdot \mid s,a) V_{h+1}^k}^2\\ 
    & \leq \frac{2H}{\gamma}\br*{\hat c_h^k(s,a) + \hat p_h^k (\cdot \mid s,a) V_{h+1}^k }\\ 
    & = \frac{2H}{\gamma} Q_h^k(s,a).
    \end{align*}
Note that the second relation holds by $(a+b)^2 \leq 2a^2 + 2b^2$. The third relation is by the fact that both terms are bounded by $\frac{H}{\gamma}$. The fourth relation is by the definition of the update rule.

Plugging this into~\eqref{eq:term 2 main term } we get for any $s\in \mathcal{S}, h\in [H]$
\begin{align*}
    &\sum_{k}\inner*{ Q_h^{k}(s,\cdot),\pi_h^k(\cdot\mid s) - \pi_h(\cdot\mid s)} \nonumber\\
     &\leq \frac{\log A}{t_K} + \frac{Ht_K}{\gamma} \sum_{k=1}^K\sum_a \pi_h^k(a \mid s) Q_h^k(s,a)\\
     & =\frac{\log A}{t_K} + \frac{Ht_K}{\gamma} \sum_{k} V_h^k(s)\\
     & \leq \frac{\log A}{t_K} + \frac{H^2t_k}{\gamma^2} \ln \frac{H^2 S}{\delta} +  \frac{Ht_K}{\gamma} \sum_{k} V_h^{\pi_k}(s)\\
     & \leq \frac{\log A}{t_K} + \frac{H^2 t_k}{\gamma^2} \ln \frac{H^2S}{\delta} +  \frac{2H^2t_K K}{\gamma}.
\end{align*}
The second relation holds by definition. The third relation holds by conditioning on the good event, specifically, event $F^{v,MD}$. The fourth relation holds since the value function of the true MDP is bounded by $H$.

Thus,
\begin{align*}
    \text{Term (ii)} &= \sum_{k=1}^K  \sum_{h=1}^H \E \brs*{\inner*{ Q_h^k(s_h,\cdot),\pi_h^k(\cdot\mid s_h) - \pi_h(\cdot\mid s_h)} \mid s_1 = s, \pi, p} \\
    &=   \sum_{h=1}^H \E \brs*{ \sum_{k=1}^K\inner*{ Q_h^k(s_h,\cdot),\pi_h^k(\cdot\mid s_h) - \pi_h(\cdot\mid s_h)} \mid s_1 = s, \pi, p} \\
    &\leq \tilde O\br*{\frac{H\log A}{t_K} + \frac{t_K H^3 }{\gamma^2} + \frac{t_K H^3 K }{\gamma}}.
\end{align*}

\begin{lemma}[Optimism Term of the Adversarial Case]\label{lemma: term 3 adversarial}
Conditioned on the good event, for any $\pi$,
\begin{align*}
    \text{Term (\romannumeral 3)}= \sum_{k=1}^K \sum_{h=1}^H \E   \brs*{Q_h^k(s_h,a_h) - c_h(s_h,a_h) - p_h(\cdot \mid s_h,a_h) V_{h+1}^{k} \mid s_1 = s,\pi, p} \leq \tilde O \br*{ \frac{H}{\gamma}} .
\end{align*}
\end{lemma}

\begin{proof}
We have that
\begin{align}
    \text{Term (iii)} &=  \sum_{k=1}^K \sum_{h=1}^H \E   \brs*{Q_h^k(s_h,a_h) - c_h(s_h,a_h) - p_h(\cdot \mid s_h,a_h) V_{h+1}^{k} \mid s_1 = s,\pi, p} \\
    & = \underbrace{\sum_{k=1}^K \sum_{h=1}^H \E   \brs*{\hat  c^k_h(s_h,a_h) -c^k_h(s_h,a_h)\mid s_1 = s,\pi, p}}_{(A)} \nonumber \\
    &+ \underbrace{\sum_{k=1}^K \sum_{h=1}^H\E  \brc*{ \hat p_h^k(\cdot \mid s_h,a_h)V_{h+1}^k  -  p_h(\cdot\mid s_h,a_h)V_{h+1}^k  \mid s_1 = s,\pi, p}}_{(B)}. \label{eq: term 3}
\end{align}

We shall prove that, conditioned on the good event,
\begin{align*}
    \text{Term (iii)} \leq \frac{H }{\gamma}\ln\frac{SAH}{\delta'}.
\end{align*}

\paragraph{Term (A).} We have that for any $s,a,h$, conditioning on the good event
\begin{align*}
    \sum_{k} \hat  c^k_h(s,a)-c^k_h(s,a)  \leq  \sum_{k} \hat  c^k_h(s,a)- \frac{d_h^k(s)}{u_h^k(s)}c^k_h(s,a) \le \frac{1}{2\gamma}\ln\frac{SAH}{\delta'},
\end{align*}
where we used that fact that conditioned on the good event, $0 \leq d_h^k(s)<u_h^k(s)$ for any $k,h,s$.

Thus,
\begin{align*}
    \text{Term (A)} &= \sum_{k=1}^K \sum_{h=1}^H \E   \brs*{\hat  c^k_h(s_h,a_h) -c^k_h(s_h,a_h)\mid s_1 = s,\pi, p}\\
    & =  \sum_{h=1}^H \E   \brs*{\sum_{k=1}^K \hat  c^k_h(s_h,a_h) -c^k_h(s_h,a_h) \mid s_1 = s,\pi, p} \leq \frac{H}{2\gamma}\ln\frac{SAH}{\delta'}.
\end{align*}

\paragraph{Term (B).} 
For any $k,h,s,a$, 
\begin{align*}
    \hat p_h^{k}(\cdot \mid s,a) V_{h+1}^k  -  p_h(\cdot\mid s,a) V_{h+1}^k = \min_{ \hat p(\cdot \mid s,a)\in  \mathcal{P}_h^{k-1}(s,a)} \hat p(\cdot \mid s,a) V^{k}_{h+1}  -  p_h(\cdot\mid s,s)V_{h+1}^k \leq 0,
\end{align*}
since $p_h(\cdot\mid s,a)\in \mathcal{P}_h^{k}(s,a)$ conditioning on the good event.

The result follows by combining the two above terms

\end{proof}

\section{Difference Lemmas}\label{sec: difference lemmas}

The following lemma is similar to the analysis of the first term, in \citep{cai2019provably}[Lemma 4.2].
\lemmaExtendedValueDiff*

\begin{proof}

For any two policies $\pi, \pi'$, and for any $h$ and $s$, by the definition $\hat V_h^{\pi,\M}(s) = \inner*{\hat Q_h^{\pi,\M}(s,\cdot) ,\pi_h(\cdot\mid s)}$ and by the definition of $V_h^{\pi',\M'},Q_h^{\pi',\M'}$,

\begin{align*}
\hat V_h^{\pi,\M}&(s) - V_h^{\pi',\M'}(s) = \inner*{\hat Q_h^{\pi,\M \phantom{'}}(s,\cdot), \pi_h(\cdot \mid s )} - \inner*{Q_h^{\pi',\M'}(s,\cdot), \pi'_h(\cdot \mid s )}\\
& = \inner*{\hat Q_h^{\pi,\M \phantom{'}}(s,\cdot), \pi_h(\cdot \mid s )-\pi'_h(\cdot \mid s )} + \inner*{\hat Q_h^{\pi,\M}(s,\cdot) - Q_h^{\pi',\M'}(s,\cdot), \pi'_h(\cdot \mid s )} \\
& = \inner*{\hat Q_h^{\pi,\M \phantom{'}}(s,\cdot), \pi_h(\cdot \mid s )-\pi'_h(\cdot \mid s )} \\
& + \inner*{\hat Q_h^{\pi,\M \phantom{'}}(s,\cdot) - c'_h(s,\cdot) - \sum_{s'} p'_h(s'\mid s,\cdot) V_{h+1}^{\pi',\M'}(s') , \pi'_h(\cdot \mid s )},
\end{align*}
where in the last relation we used the fixed-policy Bellman equation on the MDP $\M'$. I.e., for any $s,a$, we have that $Q_h^{\pi',\M'}(s,a) = c'_h(s,a) + \sum_{s'} p'_h(s'\mid s,a) V_{h+1}^{\pi',\M'}(s')$.

Now, by adding and subtracting $\sum_{s'}  p'_h(s'\mid s,\cdot) \br*{\hat V_{h+1}^{\pi,\M}(s') , \pi'_h(\cdot \mid s )}$, we get
\begin{align*}
\hat V_h^{\pi,\M}&(s) - V_h^{\pi',\M'}(s) = \inner*{\hat Q_h^{\pi,\M \phantom{'}}(s,\cdot), \pi_h(\cdot \mid s )-\pi'_h(\cdot \mid s )} \\
& + \inner*{\hat Q_h^{\pi,\M \phantom{'}}(s,\cdot) - c'_h(s,\cdot) - \sum_{s'} p'_h(s'\mid s,\cdot) \hat V_{h+1}^{\pi,\M}(s') , \pi'_h(\cdot \mid s )} \\
& + \inner*{\sum_{s'}  p'_h(s'\mid s,\cdot) \br*{\hat V_{h+1}^{\pi,\M}(s') - V_{h+1}^{\pi',\M'}(s')} , \pi'_h(\cdot \mid s )} \\ 
& = \inner*{\hat Q_h^{\pi,\M \phantom{'}}(s,\cdot), \pi_h(\cdot \mid s )-\pi'_h(\cdot \mid s )} \\
& + \sum_a \pi'_h(a\mid s) \br*{\hat Q_h^{\pi,\M \phantom{'}}(s,a) - c'_h(s,a) - \sum_{s'} p'_h(s'\mid s,a) \hat V_{h+1}^{\pi,\M}(s')} \\
& + \sum_{s'}\sum_a  p'_h(s'\mid s,a)\pi'_h(a \mid s ) \br*{\hat V_{h+1}^{\pi,\M}(s') - V_{h+1}^{\pi',\M'}(s')} \\
& = \inner*{\hat Q_h^{\pi,\M \phantom{'}}(s,\cdot), \pi_h(\cdot \mid s )-\pi'_h(\cdot \mid s )} \\
& + \E \brs*{\hat Q_h^{\pi,\M \phantom{'}}(s,a) - c'_h(s,a) - \sum_{s'} p'_h(s'\mid s,a) \hat V_{h+1}^{\pi,\M}(s') \mid s_h = s,\pi',\M' } \\
& + \E \brs*{\hat V_{h+1}^{\pi,\M}(_{h+1}) - V_{h+1}^{\pi',\M'}(s_{h+1})\mid s_h = s,\pi',\M' } \\
\end{align*}

By using the above relation recursively, we obtain,
\begin{align*}
&\hat V_1^{\pi,\M}(s) - V_1^{\pi',\M'}(s)\\
&=\E  \sum_{h=1}^H \brs*{ \inner*{\hat Q_h^{\pi,\M\phantom{'}}(s_h,\cdot), \pi_h (\cdot \mid s_h )-\pi'_h(\cdot \mid s_h )}\mid s_1 = s,\pi',\M'} \\
& + \E \sum_{h=1}^H \brs*{\hat Q_h^{\pi,\M \phantom{'}}(s_h,a_h) - c'_h(s_h,a_h) - \sum_{s'} p'_h(s'\mid s_h,a_h) \hat V_{h+1}^{\pi,\M}(s')\mid s_h=s,\pi', \M'}\\
& +  \E\brs*{\hat V_{H+1}^{\pi,\M}(s_{H+1}) - V_{H+1}^{\pi',\M'}(s_{H+1}) \mid s_1 = s, \pi', \M'} .
\end{align*}

By using the fact that for any policy $H$-horizon MDP $\M$ and for any policy $\pi$ and state $s$, $\hat V_{H+1}^{\pi,\M}(s)=0$, we get
\begin{align*}
& \hat V_1^{\pi,\M}(s) - V_1^{\pi',\M'}(s)\\
& = \sum_{h=1}^H \E \brs*{ \inner*{Q_h^{\pi\phantom{'}}(s_h,\cdot), \pi_h(\cdot \mid s_h )-\pi'_h(\cdot \mid s_h )} \mid s_1 = s,\pi',\M'}\\
&+ \sum_{h=1}^H\E   \brs*{\hat Q_{h}^{\pi,\M}(s_h) - c'_h(s_h,a_h) -  p'_h(\cdot \mid s_h,a_h) \hat V_{h+1}^{\pi,\M}\mid s_1 = s,\pi', \M'} ,
\end{align*}
\end{proof}
which concludes the proof.

By replacing the approximation in the last lemma with the real expected value, we get the following well known result: 

\begin{corollary}[Value difference]\label{corollary: value difference}
Let $\M,\M'$ be any $H$-finite horizon MDP.
Then, for any two policies $\pi,\pi'$, the following holds
\begin{align*}
    & V_1^{\pi,\M}(s) - V_1^{\pi',\M'}(s)
    =
    \\
    & \quad =
    \sum_{h=1}^H \E \brs*{ \inner*{ Q_h^{\pi, \M\phantom{'}}(s_h,\cdot), \pi_h(\cdot \mid s_h )-\pi'_h(\cdot \mid s_h )} \mid s_1 = s,\pi',\M'}
    \\
    & \quad + 
    \sum_{h=1}^H \E \brs*{ \br*{c_h(s_h,a_h) - c'_h(s_h,a_h)} + \br*{p_h(\cdot \mid s_h,a_h)- p'_h(\cdot \mid s_h,a_h)} V_{h+1}^{\pi,\M} \mid s_h=s,\pi', \M'} .
\end{align*}
\end{corollary}

\section{Useful Lemmas}\label{sec: useful lemmas}

\subsection{Online Mirror Descent}

In each iteration of Online Mirror Descent (OMD), the following problem is solved:

\begin{align}\label{eq: OMD iterates}
x_{k+1} \in \argmin_{x\in \Delta_d} t_K \inner*{g_k, x - x_k } + \bregman{x}{x_k}.
\end{align}

The following lemma, \cite{orabona2019modern}[Theorem 10.4],  provides a fundamental inequality which will be used in our analysis.
\begin{lemma}[Fundamental inequality of Online Mirror Descent, \citealt{orabona2019modern}, Theorem 10.4]\label{lemma: OMD orabona}
Assume for $g_{k,i} \geq 0$ for $k=1,...,K$ and $i=1,...,d$. Let $C = \Delta_d$ and $\eta>0$. Using OMD with the KL-divergence, learning rate $t_K$, and with uniform initialization, $x_1=[1/d,...,1/d]$, the following holds for any $u\in \Delta_d$,
$$ \sum_{k=1}^K \inner*{ g_t, x_k - u } \leq \frac{\log d}{t_K} + \frac{t_K}{2} \sum_{k=1}^K \sum_{i=1}^d x_{k,i} g_{k,i}^2 $$
\end{lemma}

In our analysis, we will be solving the OMD problem for each time-step $h$ and state $s$ separately,
\begin{align}\label{eq: RL OMD iterates}
     \pi_h^{k+1}(\cdot \mid s) \in \argmin_{\pi \in \simplex} t_K \inner*{Q_h^k(s,\cdot), \pi - x_h^k(\cdot \mid s) } + \dkl{\pi}{\pi_h^k(\cdot\mid s)}.
\end{align}

Therefore, by adapting the above lemma to our notation, we get the following lemma,

\begin{lemma}[Fundamental inequality of Online Mirror Descent for RL]\label{lemma: fundamental inequality of OMD}
Let $t_K>0$. Let $\pi_h^1(\cdot \mid s)$ be the uniform distribution for any $h\in[H]$ and $s\in \sset$. Then, by solving \eqref{eq: RL OMD iterates} separately for any $k\in[K], h\in[H]$ and $s\in \sset$, the following holds for any stationary policy $\pi$,
$$\sum_{k=1}^K \inner*{ Q_h^k( \cdot \mid  s), \pi_h^k(\cdot \mid s) - \pi_h(\cdot \mid s) } \leq \frac{\log A}{t_K} + \frac{t_K}{2} \sum_{k=1}^K \sum_a \pi_h^k(a \mid s) (Q_h^k(s,a))^2 $$
\end{lemma}
\begin{proof}
First, observe that for any $k,h,s$, we solve the optimization problem defined in \eqref{eq: RL OMD iterates} which is the same as \eqref{eq: OMD iterates}. By the fact that the estimators used in our analysis are non-negative, we can apply Lemma~\ref{lemma: OMD orabona} separately for each $h,s$ with $g_k = Q_h^k(s,\cdot)$ and $x_k = \pi_h^k(s,\cdot) $.
\end{proof}

\subsection{Bounds on the Visitation Counts}

\begin{lemma}[e.g.\ \citealt{zanette2019tighter}, Lemma 13]
\label{lemma: supp 1 1/N  factor and lograthimic factors}
Outside the failure event, it holds that
$$\sum_{k=1}^K\sum_{t=1}^H \E\brs*{ {\frac{1}{n_{k-1}(s_t^k,\pi_k(s_t^k))\vee 1}} \mid \F_{k-1} }\leq \Tilde{O}\br*{SAH^2}.$$
\end{lemma}

\begin{lemma}[e.g.\ \citealt{efroni2019tight}, Lemma 38]
\label{lemma: supp 1 factor and lograthimic factors}
Outside the failure event, it holds that
$$\sum_{k=1}^K\sum_{t=1}^H \E\brs*{ \sqrt{\frac{1}{n_{k-1}(s_t^k,\pi_k(s_t^k))\vee 1}} \mid \F_{k-1} }\leq \Tilde{O}\br*{\sqrt{SAH^2K} +SAH}.$$
\end{lemma}

In both~\citealt{zanette2019tighter,efroni2019tight} these results were derived for MDPs with stationary dynamics. Repeating their analysis, in our case, an additional $H$ factor emerges as we consider MDPs with non-stationary dynamics.

\subsection{Bias Lemmas}

\lemmaBiasAdversarialCosts*
\begin{proof}
For any $k$ and state-action pair $(s,a)$ we have
\begin{align}
    \nonumber
    \hat c_h^k (s,a)
    & =
    \frac{ c_h^k(s,a) \I \br*{s_h^k=s,a_h^k=a} }{ u_h^k(s)\pi_h^k(a \mid s) + \gamma}
    \\
    \nonumber
    & \le
    \frac{ c_h^k(s,a) \I \br*{s_h^k=s,a_h^k=a} }{ u_h^k(s)\pi_h^k(a \mid s) + \gamma c_h^k(s,a)}
    \\
    \nonumber
    & =
    \frac{\I \br*{s_h^k=s,a_h^k=a}}{2 \gamma} \cdot \frac{2 \gamma c_h^k(s,a)}{u_h^k(s)\pi_h^k(a \mid s) + \gamma c_h^k(s,a)}
    \\
    \nonumber
    & =
    \frac{\I \br*{s_h^k=s,a_h^k=a}}{2 \gamma} \cdot \frac{2 \gamma c_h^k(s,a) / (u_h^k(s)\pi_h^k(a \mid s))}{1 + \gamma c_h^k(s,a) / (u_h^k(s)\pi_h^k(a \mid s))}
    \\
    \nonumber
    & \le
    \frac{\I \br*{s_h^k=s,a_h^k=a}}{2 \gamma} \ln \Biggl( 1 + \frac{2 \gamma c_h^k(s,a)}{u_h^k(s)\pi_h^k(a \mid s)} \Biggr)
    \tag{$\frac{z}{1 + z/2} \le \ln (1 + z)$ for $z \ge 0$}
    \\
    \label{eq:bound-hat-c-with-ln}
    & =
    \frac{1}{2 \gamma} \ln \Biggl( 1 + \frac{2 \gamma c_h^k(s,a) \I \br*{s_h^k=s,a_h^k=a} }{u_h^k(s)\pi_h^k(a \mid s)} \Biggr).
\end{align}

Define $\hat X_h^k = \sum_{s,a} \alpha_h^k(s,a) \hat c_h^k(s,a)$ and $X_h^k = \sum_{s,a} \alpha_h^k(s,a) \frac{  d_h^k(s) }{u_h^k(s)} c_h^k(s,a)$.
Next, we prove that $\E \brs*{ \exp ( \hat X_h^k ) \mid \filt } \le \exp ( X_h^k )$.
\begin{align*}
    \E \brs*{ \exp ( \hat X_h^k ) \mid \filt }
    & =
    \E \brs*{ \exp \Biggl( \sum_{s,a} \alpha_h^k (s,a) \hat c_h^k (s,a) \Biggr) \mid \filt }
    \\
    & \le
    \E \brs*{ \exp \Biggl( \sum_{s,a} \frac{\alpha_h^k (s,a)}{2 \gamma} \ln \Biggl( 1 + \frac{2 \gamma c_h^k(s,a) \I \br*{s_h^k=s,a_h^k=a} }{u_h^k(s)\pi_h^k(a \mid s)} \Biggr) \Biggr)  \mid \filt }
    \tag{by eq. \eqref{eq:bound-hat-c-with-ln}}
    \\
    & \le
    \E \brs*{ \exp \Biggl( \sum_{s,a} \ln \Biggl( 1 + \frac{ \alpha_h^k(s,a) c_h^k(s,a) \I \br*{s_h^k=s,a_h^k=a} }{u_h^k(s)\pi_h^k(a \mid s)} \Biggr) \Biggr)  \mid \filt }
    \tag{$z_1 \ln (1 + z_2) \le \ln (1 + z_1 z_2)$ for $z_1 \in [0,1]$, $z_2 \ge -1$}
    \\
    & =
    \E \brs*{ \prod_{s,a} \Biggl( 1 + \frac{ \alpha_h^k(s,a) c_h^k(s,a) \I \br*{s_h^k=s,a_h^k=a} }{u_h^k(s)\pi_h^k(a \mid s)} \Biggr)  \mid \filt }
    \\
    & =
    \E \brs*{ 1 + \sum_{s,a} \frac{ \alpha_h^k(s,a) c_h^k(s,a) \I \br*{s_h^k=s,a_h^k=a} }{u_h^k(s)\pi_h^k(a \mid s)}  \mid \filt }
    \tag{indicator is zero for all but one state-action pair}
    \\
    & =
    1 + \sum_{s,a} \alpha_h^k(s,a) \frac{  d_h^k(s) }{u_h^k(s)} c_h^k(s,a)
    =
    1 + X_h^k
    \le
    \exp ( X_h^k ).
\end{align*}

Now, we use the above relation and apply Markov inequality to obtain
\begin{align*}
    \Pr \Biggl[ \sum_{k=1}^K \hat X_h^k - X_h^k > \ln \frac{H}{\delta} \Biggr] 
    & =
    \Pr \Biggl[ \exp \Biggl( \sum_{k=1}^K \hat X_h^k - X_h^k \Biggr) > \frac{H}{\delta} \Biggr]
    \\
    & \le
    \frac{\delta}{H} \E \brs*{ \exp \Biggl( \sum_{k=1}^K \hat X_h^k - X_h^k \Biggr) }
    \\
    & = 
    \frac{\delta}{H} \E \brs*{ \exp \Biggl( \sum_{k=1}^{K-1} \hat X_h^k - X_h^k \Biggr) \E \brs*{ \exp \Biggl( \hat X_h^K - X_h^K \Biggr) \mid \mathcal{F}_K } }
    \\
    & \le 
    \frac{\delta}{H} \E \brs*{ \exp \Biggl( \sum_{k=1}^{K-1} \hat X_h^k - X_h^k \Biggr) }
    \le
    \ldots
    \le
    \frac{\delta}{H},
\end{align*}
where the last inequality follows because $\E \brs*{ \exp ( \hat X_h^k ) \mid \filt } \le \exp ( X_h^k )$.
\end{proof}

\begin{lemma}[\citealt{jin2019learning}, Lemma 4]
\label{lemma: occupancy-measures-distance}
For any $k$, let $\{ \tilde{p}^{k,s} \}_{s \in \sset}$ be any collection of transition functions which are all $\filt$-measurable and belong to $\P^k$.
Define the visitation frequencies
\begin{align*}
    d_h^k(s) 
    & = 
    \E \brs*{ \I \br*{s_h^k=s} \mid \pi^k, p }
    \\
    \tilde{d}_h^{k,s}(s) 
    & = 
    \E \brs*{ \I \br*{s_h^k=s} \mid \pi^k, \tilde{p}^{k,s} },
\end{align*}
for every $(s,h,k) \in \sset \times [H] \times [K]$.
With probability at least $1 - \delta'$, 
\[
    \sum_{k=1}^K \sum_{h=1}^H \sum_{s \in \sset} | d_h^k(s) - \tilde{d}_h^{k,s}(s)|
    \le
    O \br*{ H S \sqrt{A T \ln \frac{SAHK}{\delta'}}} .
\]

Notice that $u_h^k(s) = \tilde{d}_h^{k,s}(s)$ for some $\tilde{p}^{k,s}$ which maximizes the probability to reach $s$ in the $h$ step of episode $k$.
Thus, with probability at least $1 - \delta'$,
\[
    \sum_{k=1}^K \sum_{h=1}^H \sum_{s \in \sset} \abs{u_h^k(s) - d_h^k(s)} 
    \le
    O \br*{ H S \sqrt{A T \ln \frac{SAHK}{\delta'}}} .
\]
\end{lemma}

\end{appendices}
\end{document}